\documentclass[letterpaper, 10 pt, conference]{ieeeconf}  
\usepackage[font=small]{caption}
\usepackage[noadjust]{cite}

\usepackage{amsmath}
\usepackage{empheq}
\usepackage{multirow}
\usepackage{bm}
\usepackage{graphicx}
\usepackage{amssymb}
\usepackage{svg}
\usepackage{array}
\usepackage{url}
\usepackage{hyperref}
\usepackage{lipsum}
\usepackage[capitalise]{cleveref}
\crefformat{equation}{(#2#1#3)}
\usepackage{tikz}
\usepackage{tikz-3dplot}
\let\labelindent\relax
\usepackage{enumitem}
\tikzset{small dot/.style={fill=black,circle,outer sep=8pt,scale=0.25}}
\usetikzlibrary{decorations.text}
\usetikzlibrary{shapes.multipart}
\makeatletter

\newif\ifpgfcirclecrosssplitcustomfill
\tikzset{%
	circle cross split part fill/.code=\def\pgf@lib@sh@ccs@list@fill{#1}\pgfcirclecrosssplitcustomfilltrue,%
	circle cross split uses custom fill/.is if=pgfcirclecrosssplitcustomfill}
\pgfdeclareshape{circle cross split}{%
	\nodeparts{text,two,three,four}%
	\savedanchor\centerpoint{%
		\pgfmathsetlength\pgf@xa{\pgfkeysvalueof{/pgf/inner xsep}}%
		\pgfmathsetlength\pgf@ya{\pgfkeysvalueof{/pgf/inner ysep}}%
		\pgf@x\wd\pgfnodeparttextbox
		\pgf@yb\dp\pgfnodeparttextbox
		\pgf@yc\dp\pgfnodeparttwobox
		\ifdim\pgf@yb>\pgf@yc
		\pgf@yc\pgf@yb
		\fi
		\advance\pgf@y-\pgf@yc
		\advance\pgf@x\pgf@xa
		\advance\pgf@y-\pgf@ya
		\advance\pgf@x.5\pgflinewidth
		\advance\pgf@y-.5\pgflinewidth
	}%
	\savedanchor\twoanchor{%
		\pgfmathsetlength\pgf@xa{\pgfkeysvalueof{/pgf/inner xsep}}%
		\pgfmathsetlength\pgf@ya{\pgfkeysvalueof{/pgf/inner ysep}}%
		\advance\pgf@x.5\pgflinewidth
		\advance\pgf@x\pgf@xa
		\advance\pgf@y.5\pgflinewidth
		\advance\pgf@y\pgf@ya
		\pgf@yb\dp\pgfnodeparttextbox
		\pgf@yc\dp\pgfnodeparttwobox
		\ifdim\pgf@yb>\pgf@yc
		\pgf@yc\pgf@yb
		\fi
		\advance\pgf@y\pgf@yc
	}%
	\savedanchor\threeanchor{%
		\pgfmathsetlength\pgf@ya{\pgfkeysvalueof{/pgf/inner ysep}}%
		\pgf@x\wd\pgfnodeparttextbox
		\pgf@yb\dp\pgfnodeparttextbox
		\pgf@yc\dp\pgfnodeparttwobox
		\ifdim\pgf@yb>\pgf@yc
		\pgf@yc\pgf@yb
		\fi
		\advance\pgf@y-\pgf@yc
		\advance\pgf@y-2\pgf@ya
		\advance\pgf@y-\pgflinewidth
		\pgf@yb\ht\pgfnodepartthreebox
		\pgf@yc\ht\pgfnodepartfourbox
		\ifdim\pgf@yb>\pgf@yc
		\pgf@yc\pgf@yb
		\fi
		\advance\pgf@y-\pgf@yc
		\advance\pgf@x-\wd\pgfnodepartthreebox
	}%
	\savedanchor\fouranchor{%
		\pgfmathsetlength\pgf@xa{\pgfkeysvalueof{/pgf/inner xsep}}%
		\advance\pgf@x\wd\pgfnodepartthreebox
		\advance\pgf@x2\pgf@xa
		\advance\pgf@x\pgflinewidth
	}%
	\saveddimen\radius{%
		\pgf@y\ht\pgfnodeparttextbox
		\pgf@yb\ht\pgfnodeparttwobox
		\ifdim\pgf@yb>\pgf@y
		\pgf@y\pgf@yb
		\fi
		\pgf@yc\dp\pgfnodeparttextbox
		\pgf@yb\dp\pgfnodeparttwobox
		\ifdim\pgf@yc>\pgf@yb
		\advance\pgf@y\pgf@yc
		\else
		\advance\pgf@y\pgf@yb
		\fi
		\pgf@yb\ht\pgfnodepartthreebox
		\ifdim\pgf@yb<\ht\pgfnodepartfourbox
		\pgf@yb\ht\pgfnodepartfourbox
		\fi
		\pgf@yc\dp\pgfnodepartthreebox
		\ifdim\pgf@yc<\dp\pgfnodepartfourbox
		\advance\pgf@yb\dp\pgfnodepartfourbox
		\else
		\advance\pgf@yb\pgf@yc
		\fi
		\ifdim\pgf@yc>\pgf@y
		\pgf@y\pgf@yc
		\fi
		\pgfmathsetlength\pgf@ya{\pgfkeysvalueof{/pgf/inner ysep}}%
		\advance\pgf@y2\pgf@ya
		\pgf@x\wd\pgfnodeparttextbox
		\pgf@xa\wd\pgfnodepartthreebox
		\pgf@xb\wd\pgfnodeparttwobox
		\pgf@xc\wd\pgfnodepartfourbox
		\ifdim\pgf@xa>\pgf@x
		\pgf@x\pgf@xa
		\fi
		\ifdim\pgf@xb>\pgf@x
		\pgf@x\pgf@xb
		\fi
		\ifdim\pgf@xc>\pgf@x
		\pgf@x\pgf@xc
		\fi
		\pgfmathsetlength\pgf@xa{\pgfkeysvalueof{/pgf/inner xsep}}%
		\advance\pgf@x2\pgf@xa
		\ifdim\pgf@y>\pgf@x
		\pgf@x\pgf@y
		\fi
		\advance\pgf@x.5\pgflinewidth
		\pgfmathsetlength{\pgf@xb}{\pgfkeysvalueof{/pgf/minimum width}}%
		\pgfmathsetlength{\pgf@yb}{\pgfkeysvalueof{/pgf/minimum height}}%
		\ifdim\pgf@x<.5\pgf@xb
		\pgf@x=.5\pgf@xb
		\fi
		\ifdim\pgf@x<.5\pgf@yb
		\pgf@x=.5\pgf@yb
		\fi
		\pgfmathsetlength{\pgf@xb}{\pgfkeysvalueof{/pgf/outer xsep}}%
		\pgfmathsetlength{\pgf@yb}{\pgfkeysvalueof{/pgf/outer ysep}}%
		\ifdim\pgf@xb<\pgf@yb
		\advance\pgf@x\pgf@yb
		\else
		\advance\pgf@x\pgf@xb
		\fi
	}%
	\inheritanchorborder[from=circle]%
	\inheritanchor[from=circle]{north}%
	\inheritanchor[from=circle]{north west}%
	\inheritanchor[from=circle]{north east}%
	\inheritanchor[from=circle]{center}%
	\inheritanchor[from=circle]{west}%
	\inheritanchor[from=circle]{east}%
	\inheritanchor[from=circle]{mid}%
	\inheritanchor[from=circle]{mid west}%
	\inheritanchor[from=circle]{mid east}%
	\inheritanchor[from=circle]{base}%
	\inheritanchor[from=circle]{base west}%
	\inheritanchor[from=circle]{base east}%
	\inheritanchor[from=circle]{south}%
	\inheritanchor[from=circle]{south west}%
	\inheritanchor[from=circle]{south east}%
	\anchor{two}{\twoanchor}%
	\anchor{three}{\threeanchor}%
	\anchor{four}{\fouranchor}%
	\inheritbackgroundpath[from=circle]
	\beforebackgroundpath{%
		\pgfutil@tempdima=\radius
		\pgfmathsetlength{\pgf@xb}{\pgfkeysvalueof{/pgf/outer xsep}}%
		\pgfmathsetlength{\pgf@yb}{\pgfkeysvalueof{/pgf/outer ysep}}%
		\ifdim\pgf@xb<\pgf@yb
		\advance\pgfutil@tempdima by-\pgf@yb
		\else
		\advance\pgfutil@tempdima by-\pgf@xb
		\fi
		\advance\pgfutil@tempdima by-.5\pgflinewidth%
		\pgfsetshortenstart{0pt}%
		\pgfsetshortenend{0pt}%
		\pgfsetarrows{-}%
		\pgfpathmoveto{\pgfpointadd{\centerpoint}{\pgfqpoint{-\pgfutil@tempdima}{0pt}}}%
		\pgfpathlineto{\pgfpointadd{\centerpoint}{\pgfqpoint{\pgfutil@tempdima}{0pt}}}%
		\pgfpathmoveto{\pgfpointadd{\centerpoint}{\pgfqpoint{0pt}{-\pgfutil@tempdima}}}%
		\pgfpathlineto{\pgfpointadd{\centerpoint}{\pgfqpoint{0pt}{\pgfutil@tempdima}}}%
		\pgfusepathqstroke
	}%
	\behindbackgroundpath{%
		\pgfutil@tempdima=\radius
		\pgfmathsetlength{\pgf@xb}{\pgfkeysvalueof{/pgf/outer xsep}}%
		\pgfmathsetlength{\pgf@yb}{\pgfkeysvalueof{/pgf/outer ysep}}%
		\ifdim\pgf@xb<\pgf@yb
		\advance\pgfutil@tempdima by-\pgf@yb
		\else
		\advance\pgfutil@tempdima by-\pgf@xb
		\fi
		\advance\pgfutil@tempdima by-.5\pgflinewidth%
		\ifpgfcirclecrosssplitcustomfill%
		\pgf@lib@sh@rs@process@list{\pgf@lib@sh@ccs@list@fill}{4}%
		{%
			\pgfmathloop
			\ifnum\pgfmathcounter>4%
			\else%
			\pgf@lib@sh@getalpha\pgf@lib@sh@rs@number{\pgfmathcounter}%
			\edef\pgf@tempa{\csname pgf@lib@sh@rs@\pgf@lib@sh@rs@number @item\endcsname}%
			\ifx\pgf@tempa\pgf@lib@sh@rs@nonetext\else
			\pgfsetfillcolor{\pgf@tempa}%
			\pgf@lib@sh@ccs@angles{\pgfmathcounter}%
			\pgfpathmoveto{\centerpoint}%
			\pgfpathlineto{\pgfpointadd{\centerpoint}{\pgfqpointpolar{\pgf@lib@sh@ccs@angle}{\pgfutil@tempdima}}}%
			\pgfpatharc{\pgf@lib@sh@ccs@angle}{\pgf@lib@sh@ccs@angle@}{\pgfutil@tempdima}%
			\pgfpathclose
			\pgfusepathqfill
			\fi
			\repeatpgfmathloop
		}%
		\fi%
	}%
}
\def\pgf@lib@sh@ccs@angles#1{%
	\ifcase#1\or\def\pgf@lib@sh@ccs@angle{90}%
	\or\def\pgf@lib@sh@ccs@angle{0}%
	\or\def\pgf@lib@sh@ccs@angle{180}%
	\else\def\pgf@lib@sh@ccs@angle{270}%
	\fi
	\edef\pgf@lib@sh@ccs@angle@{\number\numexpr\pgf@lib@sh@ccs@angle+90\relax}%
}
\makeatother

\usepackage{subcaption}

\IEEEoverridecommandlockouts                              

\title{\LARGE \bf
A Collision-Free MPC for Whole-Body Dynamic Locomotion and Manipulation
}

\author{Jia-Ruei Chiu, Jean-Pierre Sleiman, Mayank Mittal, Farbod Farshidian, Marco Hutter
\thanks{This research was supported in part by the Swiss National Science Foundation through the National Centre of Competence in Research Robotics (NCCR Robotics), and in part by TenneT.}
\thanks{All authors are with the Robotic Systems Lab, ETH Zurich, Zurich 8092, Switzerland. 
M. Mittal is also with NVIDIA.
(Email: {\tt\small  jichiu@student.ethz.ch})
}
}

\newcommand{\etal}{\textit{et al}. }

\newcommand{\secref}[1]{Sec.~\ref{#1}}
\newcommand{\eqtref}[1]{Eq.~\ref{#1}}
\newcommand{\figref}[1]{Fig.~\ref{#1}}

\begin{document}

\maketitle
\thispagestyle{empty}
\pagestyle{empty}

\begin{abstract}
In this paper, we present a real-time whole-body planner for collision-free legged mobile manipulation. We enforce both self-collision and environment-collision avoidance as soft constraints within a Model Predictive Control (MPC) scheme that solves a multi-contact optimal control problem. By penalizing the signed distances among a set of representative primitive collision bodies, the robot is able to safely execute a variety of dynamic maneuvers while preventing any self-collisions. Moreover, collision-free navigation and manipulation in both static and dynamic environments are made viable through efficient queries of distances and their gradients via a euclidean signed distance field. We demonstrate through a comparative study that our approach only slightly increases the computational complexity of the MPC planning. Finally, we validate the effectiveness of our framework through a set of hardware experiments involving dynamic mobile manipulation tasks with potential collisions, such as locomotion balancing with the swinging arm, weight throwing, and autonomous door opening.
\end{abstract}

\section{INTRODUCTION} \label{Introduction}
The resemblance of legged robots to their biological counterparts has made them an ideal option for various settings such as industrial inspection, environment exploration, and search and rescue. Their unrivaled agility allows traversal over a range of rough terrains -- from a regular staircase to an unstructured subterrain. Compared to a wheel-based mobile manipulator, a legged manipulator can further exploit its six degrees-of-freedom (DoFs) floating base to increase the reachability of the arm \cite{ALMA2019}. However, due to the added complexity, motion planning and control remain highly challenging for these systems.

A common approach to the whole-body planning problem for mobile manipulation treats the base and manipulator as decoupled subsystems~\cite{PR2DoorOpening, BigDogManipulation, HyQ}. While such a decomposition yields simpler and computationally tractable problems, heuristics combining the separate plans may not suffice for large or dynamic object interaction where tight coordination of the base and arm are necessary~\cite{Mayank}. In contrast, unifying locomotion and manipulation in a single framework can generate coordinated base and arm motions. Optimal control is a tool able to manage multiple objectives and constraints jointly for the base and the manipulator~\cite{JeanPierre, Johannes, Mayank}.


\begin{figure}[t]
    \begin{subfigure}[t]{0.5\textwidth}
        \centering
        \scalebox{0.75}{
        \begin{tikzpicture}
\node (A) at (0,0) {\includegraphics[trim={29.5cm 3cm 15cm 12cm},clip,scale=0.15]{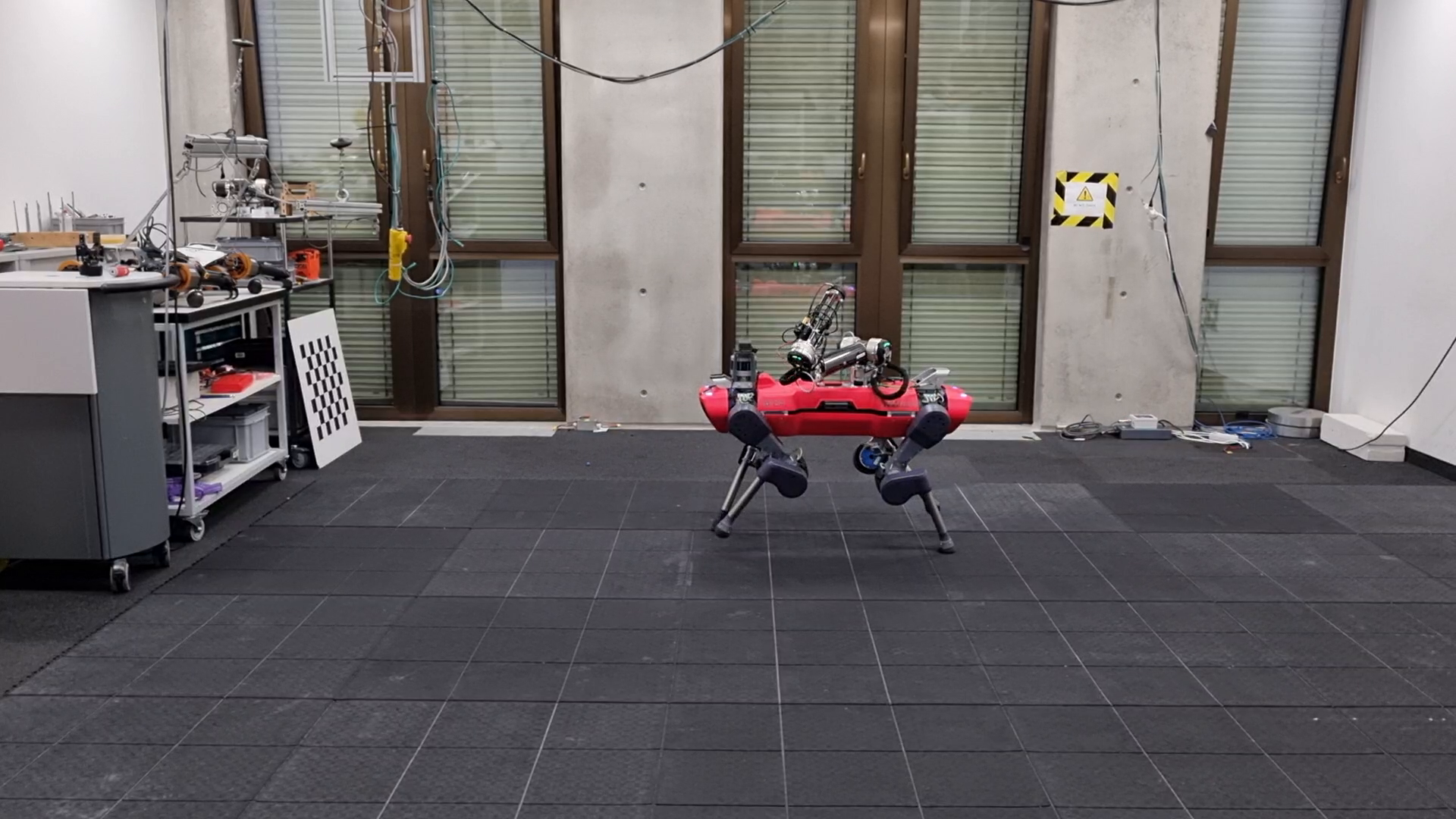}};
\node[right] (B) at (A.east) {\includegraphics[trim={29.5cm 3cm 15cm 12cm},clip,scale=0.15]{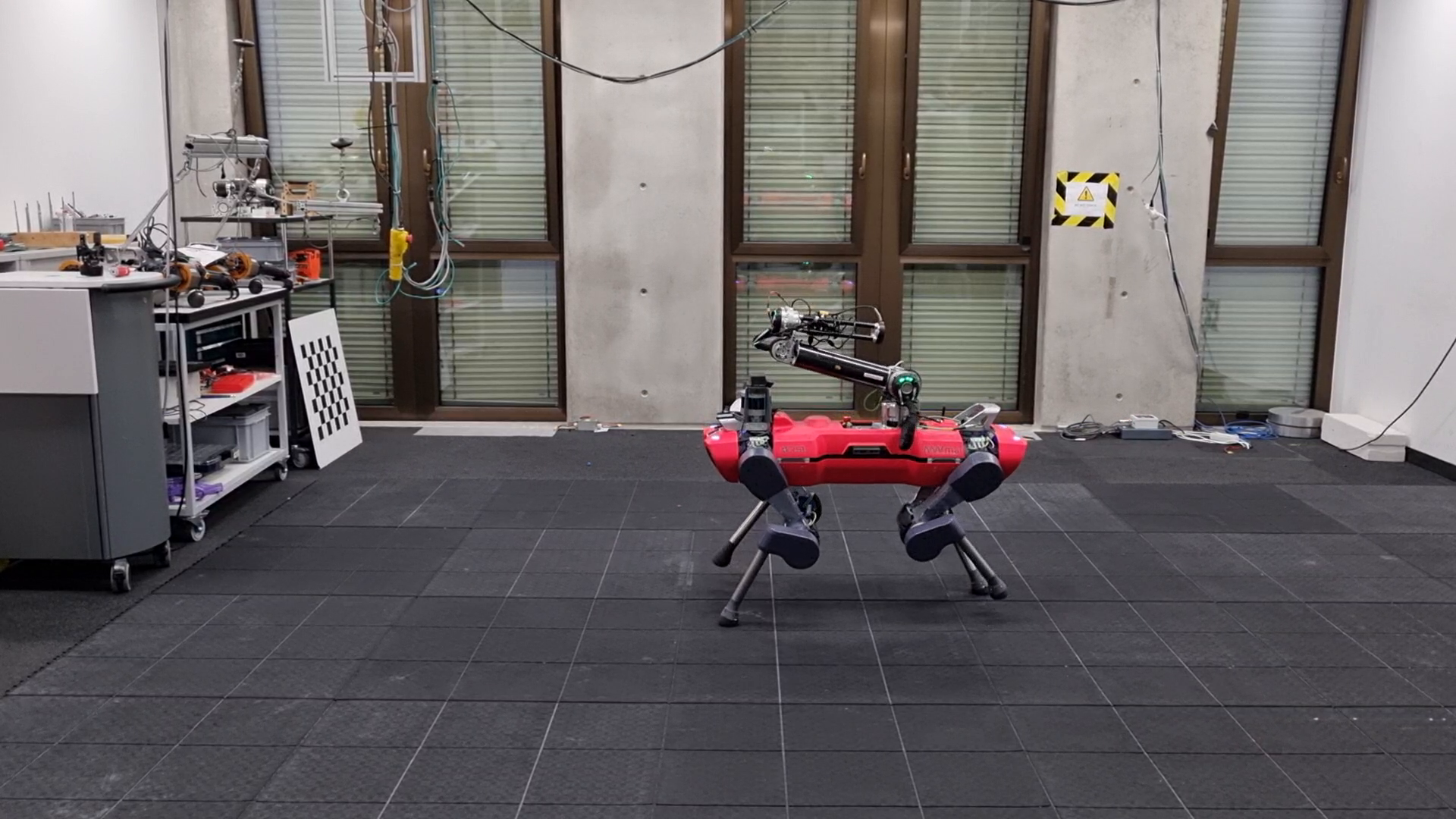}};
\node[right] (C) at (B.east) {\includegraphics[trim={29.5cm 3cm 15cm 12cm},clip,scale=0.15]{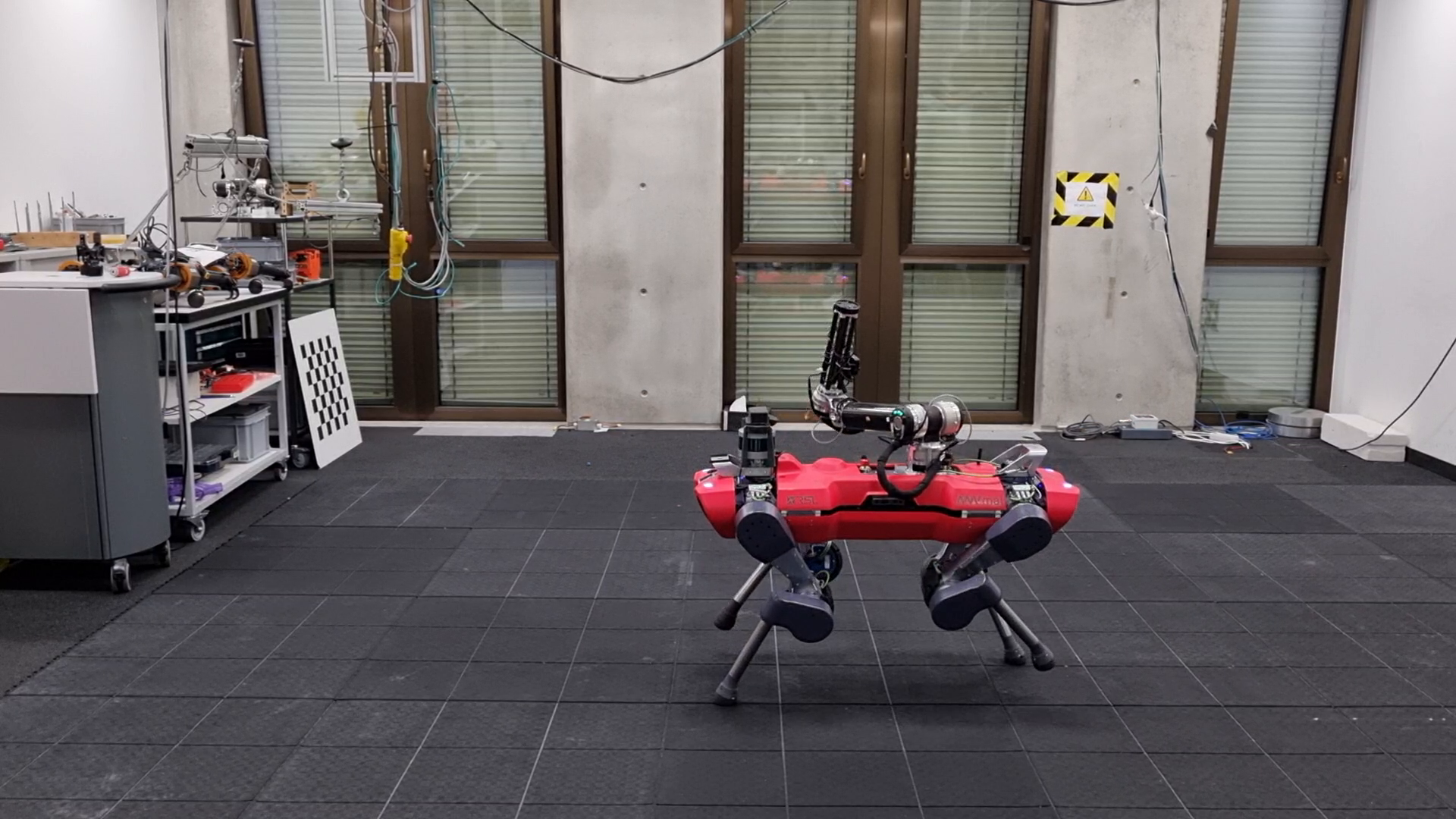}};
\node(AA) at (A.north west) {};
\node [ draw = black, minimum height = 0.3 cm, minimum width = 0.4cm, xshift = 0.675cm,yshift = -0.25cm, anchor = north east] at (AA) {\small 1};
\node(AA) at (B.north west) {};
\node [ draw = black, minimum height = 0.3 cm, minimum width = 0.4cm, xshift = 0.675cm,yshift = -0.25cm, anchor = north east] at (AA) {\small 2};
\node(AA) at (C.north west) {};
\node [ draw = black, minimum height = 0.3 cm, minimum width = 0.4cm, xshift = 0.675cm,yshift = -0.25cm, anchor = north east] at (AA) {\small 3};
\end{tikzpicture}}
        \label{fig:tail_exp_trot}
    \end{subfigure}
    \vspace{-6mm}
    
    \begin{subfigure}[t]{0.5\textwidth}
        \centering
        \scalebox{0.75}{
        \begin{tikzpicture}
\node (A) at (0,0) {\includegraphics[trim={27cm 8cm 15cm 7cm},clip,scale=0.135]{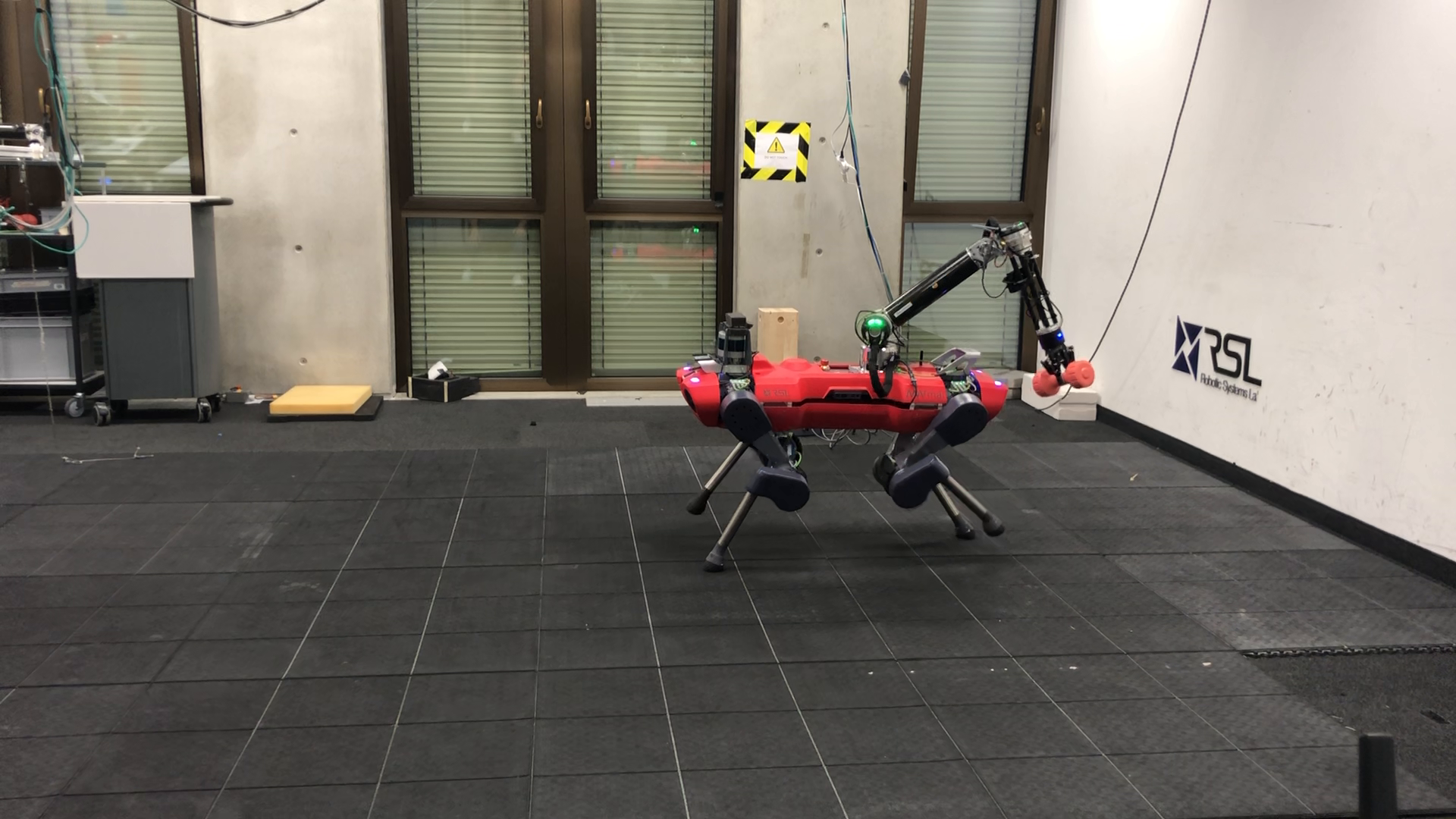}};
\node[right] (B) at (A.east) {\includegraphics[trim={27cm 8cm 15cm 7cm},clip,scale=0.135]{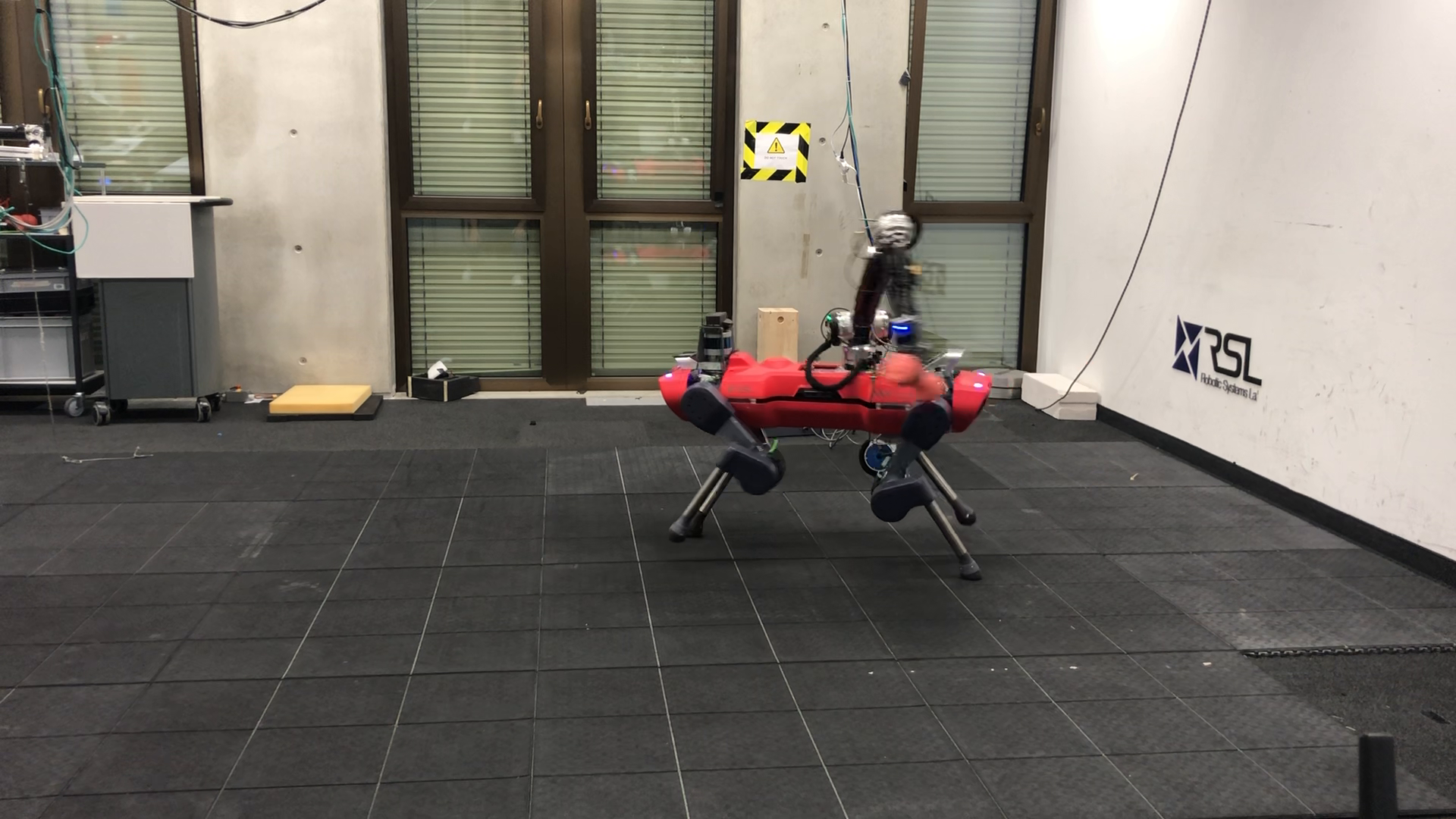}};
\node[right] (C) at (B.east) {\includegraphics[trim={27cm 8cm 15cm 7cm},clip,scale=0.135]{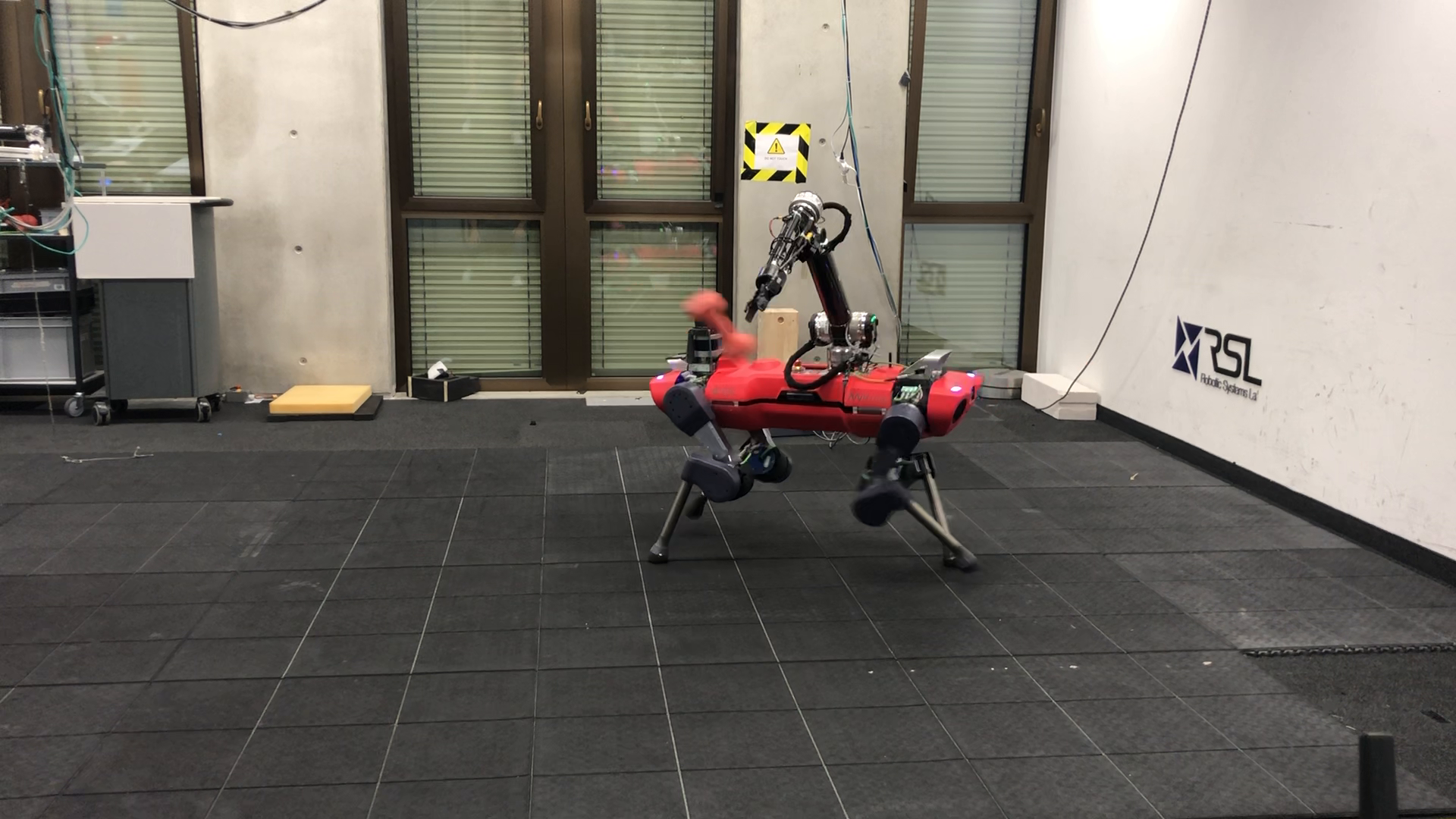}};
\node(AA) at (A.north west) {};
\node [ draw = white, minimum height = 0.3 cm, minimum width = 0.4cm, xshift = 0.675cm,yshift = -0.25cm, anchor = north east] at (AA) {\small\color{white} 4};
\node(AA) at (B.north west) {};
\node [ draw = white, minimum height = 0.3 cm, minimum width = 0.4cm, xshift = 0.675cm,yshift = -0.25cm, anchor = north east] at (AA) {\small\color{white} 5};
\node(AA) at (C.north west) {};
\node [ draw = white, minimum height = 0.3 cm, minimum width = 0.4cm, xshift = 0.675cm,yshift = -0.25cm, anchor = north east] at (AA) {\small\color{white} 6};
\end{tikzpicture}}
        \label{fig:throwing_exp_back}
    \end{subfigure}
    \vspace{-6mm}
    
    \begin{subfigure}[t]{0.5\textwidth}
        \centering
        \scalebox{0.75}{
        \begin{tikzpicture}
\node (A) at (0,0) {\includegraphics[trim={17cm 3cm 15cm 6cm},clip,scale=0.097]{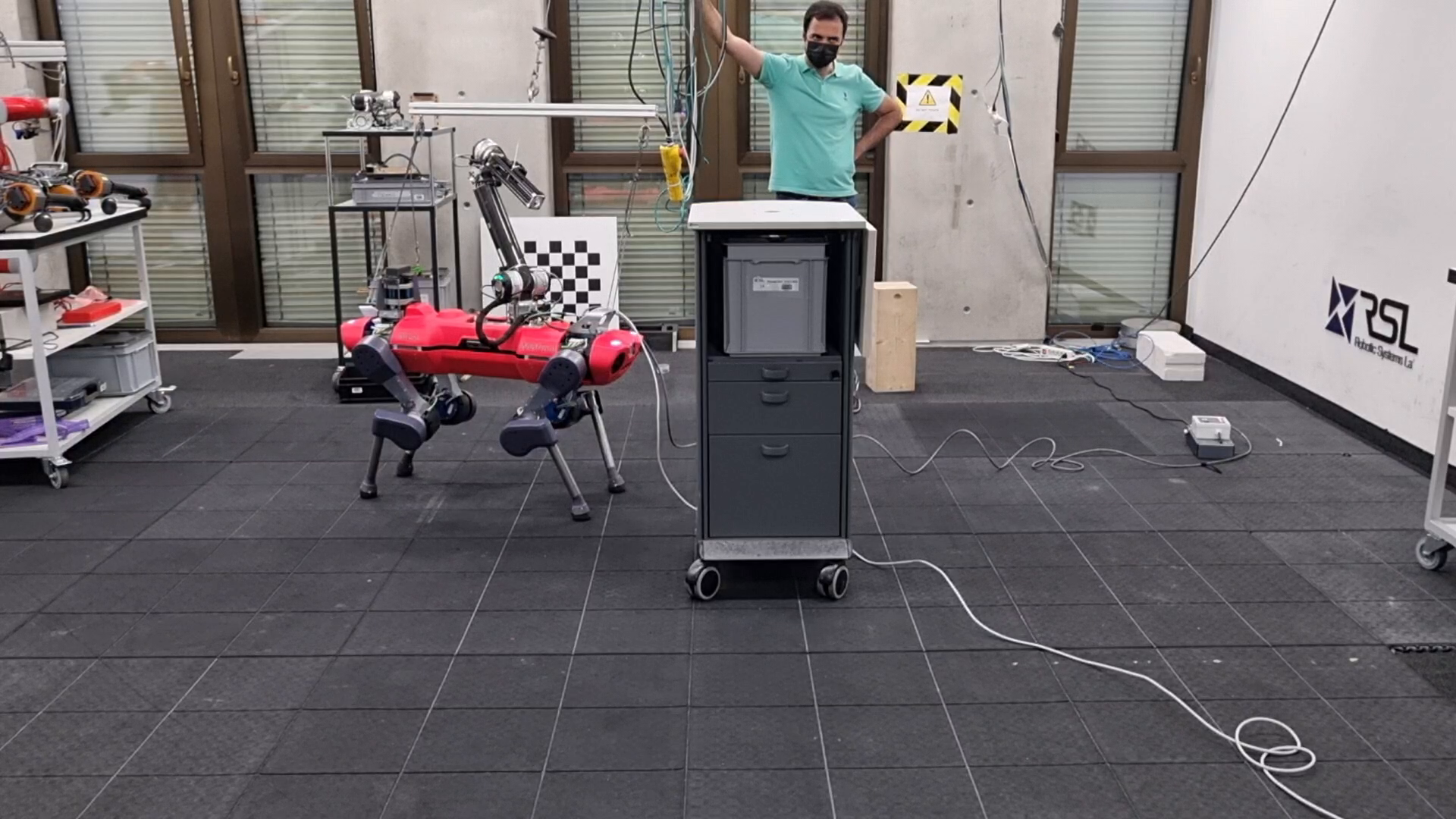}};
\node[right] (B) at (A.east) {\includegraphics[trim={17cm 3cm 15cm 6cm},clip,scale=0.097]{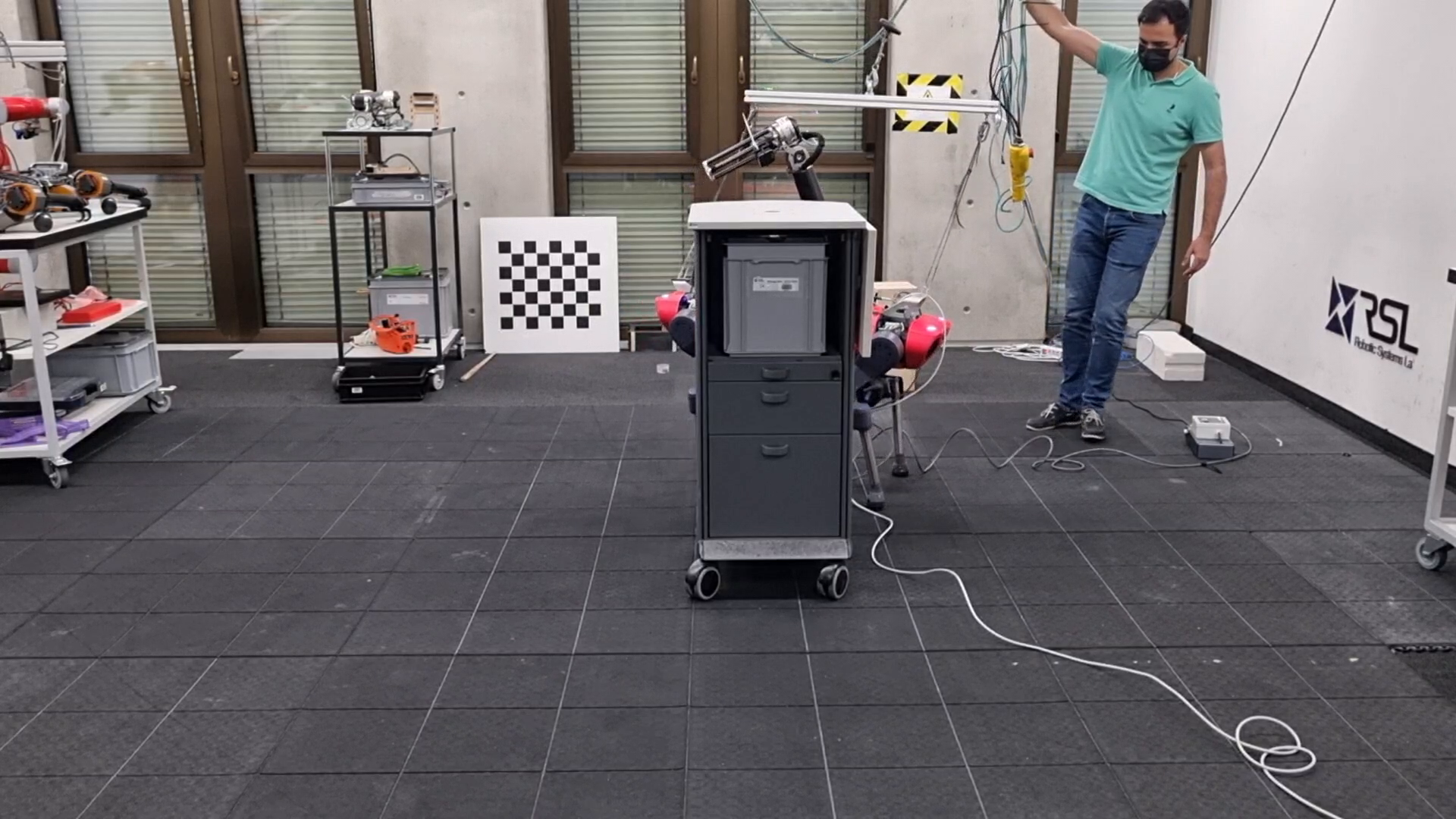}};
\node[right] (C) at (B.east) {\includegraphics[trim={17cm 3cm 15cm 6cm},clip,scale=0.097]{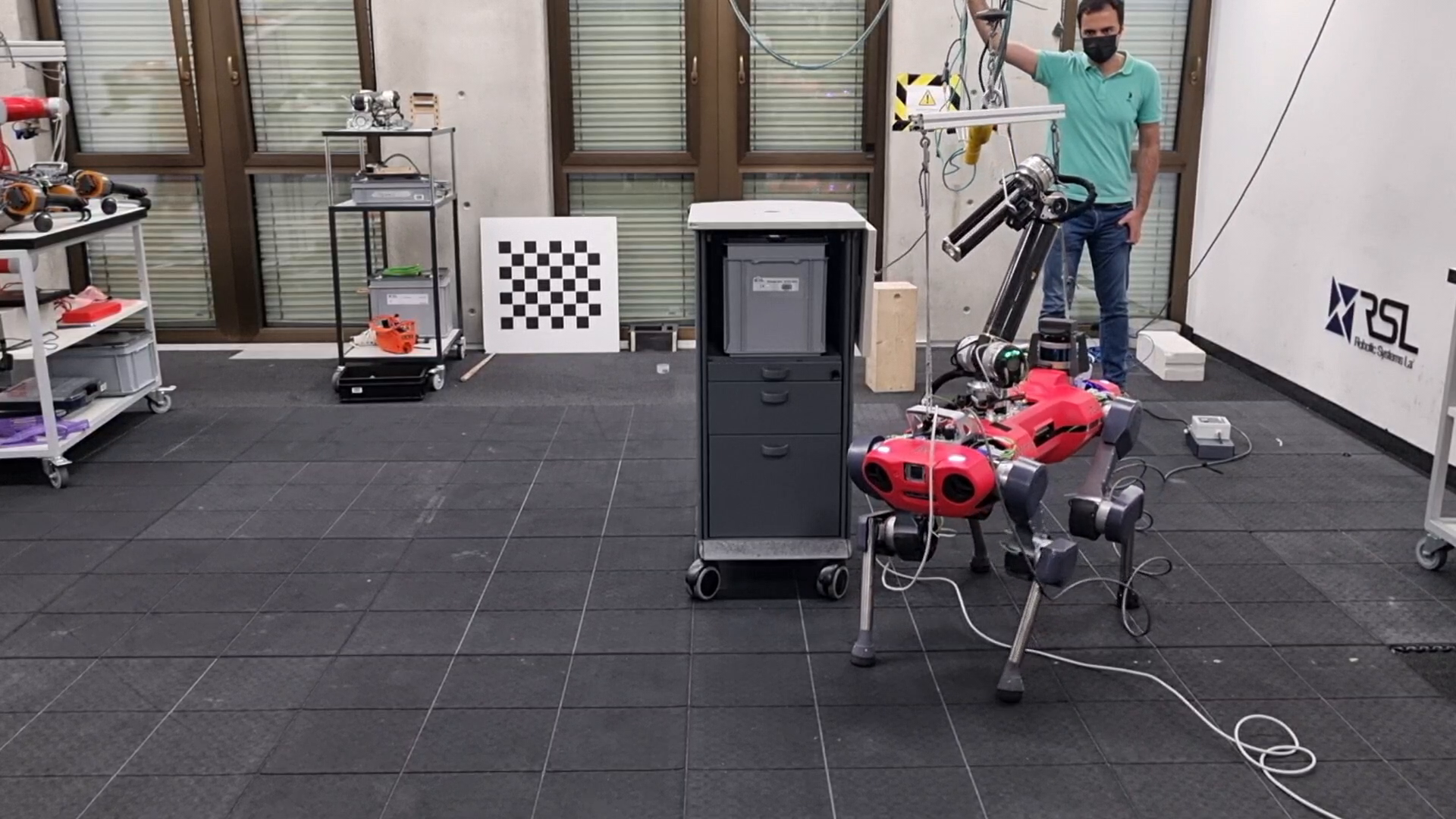}};
\node(AA) at (A.north west) {};
\node [ draw = white, minimum height = 0.3 cm, minimum width = 0.4cm, xshift = 0.675cm,yshift = -0.25cm, anchor = north east] at (AA) {\small\color{white} 7};
\node(AA) at (B.north west) {};
\node [ draw = white, minimum height = 0.3 cm, minimum width = 0.4cm, xshift = 0.675cm,yshift = -0.25cm, anchor = north east] at (AA) {\small\color{white} 8};
\node(AA) at (C.north west) {};
\node [ draw = white, minimum height = 0.3 cm, minimum width = 0.4cm, xshift = 0.675cm,yshift = -0.25cm, anchor = north east] at (AA) {\small\color{white} 9};
\end{tikzpicture}}
        \label{fig:static_cart}
    \end{subfigure}
    \caption{(1)-(3) Balancing with arm while trotting sideways and switching trotting direction without self-collision between the elbow and LiDAR cage in (2). (4)-(6) Collision-free dynamic backward weight throwing while trotting. (7)-(9) Navigation towards an end-effector position target behind a static obstacle. }
    \label{fig:teaser}
    \vspace{-6mm}
\end{figure}
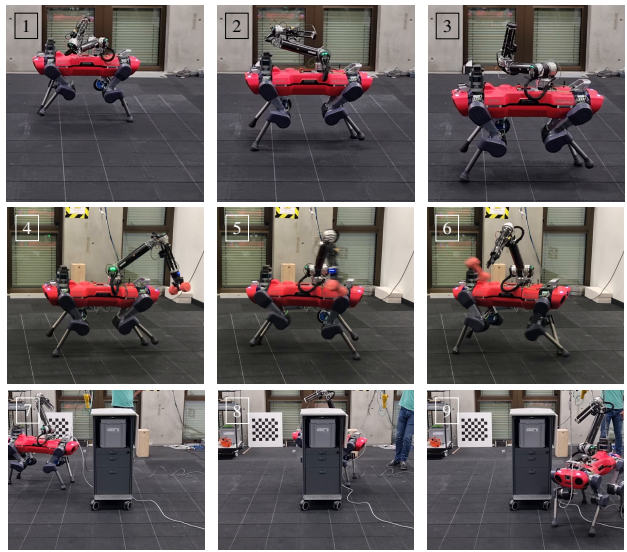

Sleiman \etal \cite{JeanPierre} propose a framework that unifies whole-body dynamic locomotion and manipulation planning into a single Model Predictive Control (MPC) problem. This formulation is applied to a legged mobile manipulator performing various free-motion and object-manipulation tasks. However, the planner does not account for any potential collisions, which restricts deployment on the robot and requires hand-crafted heuristics to prevent any damages. In this work, we build upon their formulation and extend it for considering robot self-collisions as well as environment-collisions in static and dynamic scenes. 

Our main contributions are listed as follows: 
\begin{itemize}
    \item We extend the whole-body MPC planner to avoid self- and environment-collisions during coordinated locomotion and manipulation without any added heuristics. 
    \item We present detailed comparisons and benchmarks for different collision-avoidance techniques, thereby providing a useful reference for the implementation of efficient collision-free motion planning.
    \item We enable a legged mobile manipulator to safely execute autonomous tasks on hardware. Most importantly, we show that the slight increase in computational cost does not compromise the MPC frequency, thus ensuring collision-free motions in real-time. 
\end{itemize}
To the best of our knowledge, this is the first work exhibiting dynamic and collision-free whole-body legged locomotion and manipulation on a real platform without offline planning.
\section{Related Work}
\subsection{Self-Collision Avoidance}
In recent years, learning-based methods have been explored in the context of self-collision avoidance. Support vector machines (SVM) have been widely applied to learn a continuously differentiable self-collision boundary function in joint-space \cite{JointSpaceSVM1, JointSpaceSVM2, JointSpaceSVM3}. This boundary is then used to formulate the collision constraints within a quadratic programming (QP) formulation. N{\"o}el \etal \cite{HybridCollision} apply a multi-layer perceptron to approximate the joint-space distance field to generate repulsive torques. However, these methods usually require expensive offline computation. In contrast, online approaches such as \cite{LaloSelfCollision, STPBV, SelfCollisionFVI} use the signed distances between pairs of collision primitives to provide efficient collision avoidance, but rely on reactive controllers. In this work, we adopt a similar strategy for self-collision avoidance; however, we rely on a receding-horizon planner, which provides us with both reactive as well as look-ahead capabilities.

\subsection{Environment-Collision Avoidance}
When considering motion planning problems involving collision-avoidance requirements, sampling-based algorithms have been widely studied and used for collision-free path planning~\cite{ProbabilisticRoadmap, RRTConnect}.
However, these planners often require post-processing steps to smoothen the computed trajectories and do not scale well to high-dimensional spaces.

Another prominent approach is encoding the problem as a constrained-optimization program. For instance, trajectory optimization has shown to yield smooth, locally optimal, and collision-free trajectories in cluttered environments.
A more known approach, Covariant Hamiltonian Optimization for Motion Planning (CHOMP) \cite{CHOMP1,CHOMP2}, maximizes a cost function composed of the integrals of trajectory smoothness and obstacle proximity along the path. It approximates the robot with a set of spheres to efficiently query the robot-to-obstacle distances in a pre-computed Signed Distance Field (SDF). 
Over the past few years, CHOMP has been the foundation of many motion planners. Oleynikova \etal \cite{OleynikovaContinuous, OleynikovaSafe} explore the collision-free trajectory in a similar fashion but with a Euclidean SDF (ESDF) and a smooth potential function for continuous gradients. For legged locomotion, Fankhauser \etal \cite{Fankhauser} build an SDF from a 2.5D elevation map and fit each robot foot with a sphere for collision-free swing leg trajectory planning.
Similarly, for posture control in confined spaces, Buchanan \etal \cite{BuchananHex, BuchananQuad} use CHOMP while employing two 2.5D elevation maps for the ceiling and the floor and place multiple spheres along the edges of the robot torso. 

Compared to 2.5D elevation maps, 3D volumetric maps capture the environment better for mobile manipulation. \emph{Voxblox} \cite{Voxblox} and \emph{FIESTA} \cite{FIESTA} are novel tools for volumetric ESDF mapping. Together with the use of spherical approximations of collision bodies, Pankert \etal \cite{Johannes}, and G\"{a}rtner \etal \cite{Magnus} have applied \emph{Voxblox} to avoid static obstacles for a wheel-based mobile manipulator and a legged robot respectively. By leveraging a faster update rate from \emph{FIESTA}, Mittal \etal \cite{Mayank} have shown dynamic obstacle-avoidance during object interaction with a wheel-based mobile manipulator. All of these methods enforce collision avoidance as soft constraints within an optimal control scheme.

\subsection{Sphere Decomposition of Robot}
In all the previous works, the collision spheres approximation of the robot is performed manually. In this work, we seek to automate this procedure by leveraging ideas from computer graphics. Hubbard \etal \cite{TimeCriticalSphereApprox} construct a sphere-tree with different levels of accuracy based on the approximated medial axis surface. The hierarchy of spheres is useful for broad-phase collision checking. Later works in \cite{SphereTree,AdaptiveMedialAxis} iteratively update the medial axis for a tighter sphere-tree and complete coverage. Given a user-specified number of spheres, the approach in \cite{VariationalSphere} approximates the mesh representation by minimizing the summed volume outside the object's surface. A more recent work by Voelz \etal \cite{AutomaticSphere} approximates primitive collision bodies for efficient distance and gradient queries in the ESDF. Our work implements this algorithm for decomposing the robot's collision bodies for collision avoidance.
\section{PROBLEM FORMULATION} \label{Formulation}
\subsection{Whole-Body MPC Planner} \label{WholeBodyPlanner}
As mentioned in \secref{Introduction}, our work extends the whole-body MPC planner developed in \cite{JeanPierre}. Therefore, we start by providing a brief description of the underlying framework, while also highlighting the main developments.
\subsubsection{Solver}
\label{Solver} 
The planner is based on the optimal control solver introduced in \cite{Farbod1, Farbod2} which employs the Sequential Linear Quadratic (SLQ) technique, a continuous-time variant of Differential Dynamic Programming (DDP). Extensions to the algorithm were made in \cite{Ruben1} and \cite{JeanPierre2} to further augment the formulation with inequality constraints through a relaxed-barrier method and generic constraints through an augmented-Lagrangian approach, respectively. 
\subsubsection{System Dynamics}
We apply our method to a quadrupedal mobile manipulator performing dynamic tasks. As previously shown in \cite{JeanPierre}, a suitable model for such a poly-articulated system would be a full centroidal dynamic description where the limbs of the robot are not assumed to be massless. Moreover, to properly capture the robot-object dynamic coupling, the object dynamics are augmented to the overall system flow map. Therefore, the Equations of Motion (EoM) are given by:
\begin{equation}
    \begin{cases}
        \dot{\bm p}_{com} = \sum\limits_{i = 1}^{n_c} \bm f_{c_i} + m \bm g \\
        \dot{\bm l}_{com} = \sum\limits_{i = 1}^{n_c} \bm r_{com, c_i} \times \bm f_{c_i} \\
        \dot{\bm q}_b = \bm A^{-1}_b \left(\bm h_{com} - \bm A_j \dot{\bm q}_j \right) \\   
        \dot{\bm q}_j = \bm v_j \\
        \dot{\bm q}_o = \bm v_o \\
        \dot{\bm v}_o = \bm M^{-1}_o \left(\bm J^T_{c_o} \bm f_{c_o} - \bm b_o \right)
    \end{cases}.
\end{equation}
The robot state ${\bm x_r=(\bm h_{com}, \bm q_b, \bm q_j) \in \mathbb{R}^{12 + n_a}}$ collects the centroidal momentum, base pose, and joint positions. The centroidal momentum ${\bm h_{com} = (\bm p_{com}, \bm l_{com}) \in \mathbb{R}^{6}}$ is composed of the linear and angular momentum. The input vector ${\bm u=(\bm f_{c_1},\ \dots, \bm f_{c_{n_c}}, \bm v_j) \in \mathbb{R}^{3 n_c + n_a}}$  consists of contact forces at $n_c$ contact points and joint velocities. The object state ${\bm x_o = (\bm q_o, \bm v_o) \in \mathbb{R}^{2n_o}}$ captures the object generalized positions and velocities. Furthermore, $\bm r_{com,c_i}$ is the position of the $i$-th contact point w.r.t. the center of mass, while ${\bm A(\bm q) = [\bm A_{b}(\bm q) \ \ \bm A_j(\bm q)] \in \mathbb{R}^{6 \times (6+n_a)}}$ is the centroidal momentum matrix which maps generalized velocities to centroidal momenta. 
The robot-object interaction happens through the contact force $\bm f_{c_o}$, which is mapped to generalized torques through the contact Jacobian $\bm J_{c_o}$. The term $\bm M_o$ denotes the generalized mass matrix, whereas $\bm b_o$ encapsulates the remaining generalized forces. 
\subsubsection{Cost Function} \label{CostFunction}
We encode all robot- and object-centric tasks in a single cost function. This would also include any collision-avoidance constraints that would be added as soft-constraints in the form of a penalty function as follows
\begin{align} \label{eq:CostFunction}
L(\bm x, \bm u, t)
=& \, \alpha_1 \, ||\bm r_{IE} - \bm r_{IE}^{ref}||^2_{\bm Q_{ee}} + \alpha_2 \, ||\bm x_r - \bm x_r^{ref}||^2_{\bm Q_r} \notag
\\ 
&+ \alpha_3 \,  ||\bm x_o - \bm x_o^{ref}||^2_{\bm Q_o} + ||\bm u - \bm u^{ref}||^2_{\bm R} \notag
\\ 
&+ L_c(\bm x_r, t),
\end{align}
where ${\bm r_{IE} \in \mathbb{R}^3}$ denotes the end-effector position in the inertial frame. The weighting matrices $\bm Q_{ee}, \bm Q_r$ and $\bm Q_o$ are positive semi-definite, and $\bm R$ is positive definite. The cost term $L_c(\bm x_r, t)$ represents the summation of all penalties corresponding to the self-collision and environment-collision constraints. The parameters $\alpha_1, \alpha_2, \alpha_3 \in \{0, 1\}$ are used to determine the combination of active cost terms according to the task description.
\subsubsection{Constraints}
\label{Constraints}
A common mathematical way to describe collision avoidance is by an inequality distance constraint
\begin{equation}
\label{eq:IneqDistConstraint}
    h_i(\bm x_r, t) = d_i(\bm x_r,t) - \epsilon_i \geq 0,
\end{equation}
where $d_i(\bm x,t)$ computes the distance between the $i$-th collision pair and $\epsilon_i$ denotes the minimum allowed distance threshold for each collision pair. As explained in \secref{SelfCollisionAvoidance} and \secref{EnvironmentCollisionAvoidance}, the distance function and threshold follow different definitions for self-collision and environment-collision avoidance. We recall that the SLQ-MPC solver can handle inequality path constraints as soft constraints by absorbing them into the cost function through penalty functions. In this work, we choose to penalize the collision distance constraints through relaxed barrier functions (RBF) \cite{Ruben1}:
\begin{equation}
    B(h) = 
    \begin{cases}
    -\mu\ln{(h)}, & h \geq \delta\\
    \phantom{+}\mu\beta(h; \delta), & h < \delta
    \end{cases},
\end{equation}
where $\beta(\cdot;\delta)$ is a quadratic function that yields a continuous and twice-differentiable barrier function with bounded curvature \cite{Ruben1}. As a result, \eqtref{eq:CostFunction} includes the collision avoidance cost term, $L_c(\bm x_r, t)$ in the following form
\begin{equation}
    L_c(\bm x_r, t) = \sum\limits_{i=1}^{n_p}B(h_i(\bm x_r, t)),
\end{equation}
where $n_p$ is the total number of the self-collision and environment-collision pairs.

Additional time-dependent switched constraints are included in the MPC formulation, such as contact complementarity constraints at the arm's end-effector and the feet, friction cone constraints, and input limits. The reader is referred to \cite{JeanPierre} for more details.
\subsection{Self-Collision Avoidance}
\label{SelfCollisionAvoidance}
Instead of triangular meshes, we represent the robot with primitive collision bodies and use GJK-based distance queries in \emph{FCL} \cite{FCL}. For each $i$-th collision body pair, we acquire the shortest distance $d_i$ between the nearest two points. Given the distance result and the nearest points ${}_\mathcal{I}\bm p_{i_A}$ and ${}_\mathcal{I}\bm p_{i_B}$ in the inertial frame, each self-collision distance $d_i$ in \eqtref{eq:IneqDistConstraint} can be written as the signed distance between the two nearest points as follows
\begin{align}
    d_i(\bm x_r) &= \pm ||{}_\mathcal{I}\bm p_{i_A}(\bm x_r) - {}_\mathcal{I}\bm p_{i_B}(\bm x_r)||_2
\end{align}
with a negative sign for colliding bodies and a positive sign for the non-overlapping case. For self-collision avoidance, we set the distance threshold $\epsilon_i=0.1\ m$ for all collision pairs. The gradient of the distance function is also necessary since the SLQ solver requires a Gauss-Newton Hessian approximation of the soft constraints penalties. Following the derivation in \cite{TrajOpt1,TrajOpt2}, we compute the gradient of the distance function w.r.t the robot state through
\begin{equation}
    \frac{\partial d_i}{\partial\bm x_r} = \pm\hat{\bm n}_i^T ({}_\mathcal{I}\bm J_{i_A} - {}_\mathcal{I}\bm J_{i_B}),
\end{equation}
where ${}_\mathcal{I}\bm J_{i,A}$ and ${}_\mathcal{I}\bm J_{i,A}$ are the Jacobian of ${}_\mathcal{I}\bm p_{i_A}$ and ${}_\mathcal{I}\bm p_{i_B}$ and $\hat{\bm n}_i$ is the normal vector
\begin{equation}
    \Hat{\boldsymbol{n}}_i =
    ({}_\mathcal{I}\bm p_{i_A} - {}_\mathcal{I}\bm p_{i_B}) / ||{}_\mathcal{I}\bm p_{i_A} - {}_\mathcal{I}\bm p_{i_B}||_2.
\end{equation}
\subsubsection{Narrow and Broad Phase Distance Query}
\label{NarrowAndBroad}
Given $n$ collision bodies, the na{\"i}ve approach to avoid self-collision is by performing the narrow-phase or direct distance queries between each collision pair. This purely narrow-phase method would result in the worst-case $O(n^2)$ complexity.

To reduce the complexity, we can apply the broad-phase distance query \cite{FCL}. A broad-phase manager builds a hierarchy of bounding boxes for a set of collision bodies. Thus, the distance query between an object and a manager is performed by recursive traversal and continues in-depth only if it reduces the shortest distance currently cached. In this manner, we avoid narrow-phase queries for faraway collision objects in the manager. Furthermore, each broad-phase query would only yield the distance result to the nearest managed collision body. Fewer distance results also mean less time for computing gradients. Overall, using the broad-phase strategy can alleviate the computation of proximity queries. Through thorough comparisons, we later show in~\secref{sec:benchmark} whether it is also advantageous in solving the MPC problem. 

\subsubsection{Robot Collision Modeling}
\label{RobotCollisionModeling}
\begin{figure}[t]
  \begin{minipage}[t]{0.155\textwidth}
    \includegraphics[width = \textwidth]{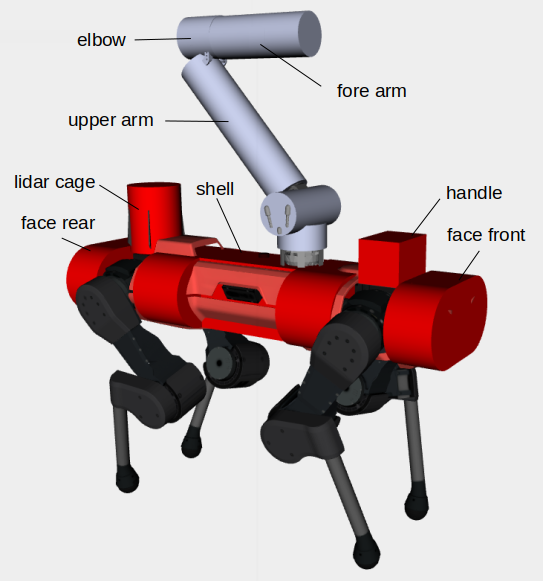}
  \end{minipage}
  \begin{minipage}[t]{0.155\textwidth}
    \includegraphics[width = \textwidth]{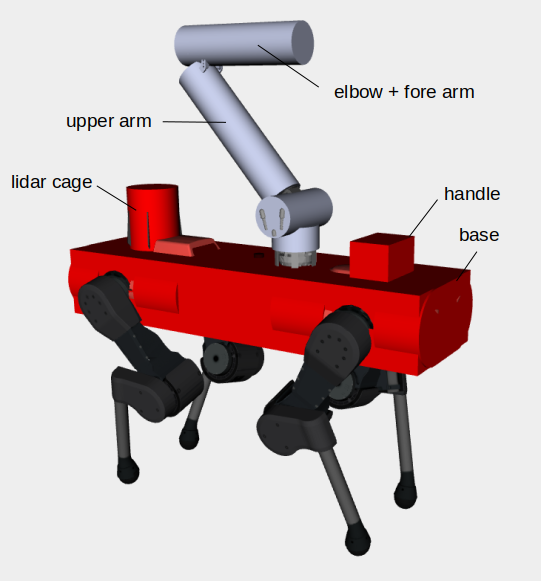}
  \end{minipage}
  \begin{minipage}[t]{0.143\textwidth}
    \includegraphics[trim=0cm 0 0cm 0,clip,width= \textwidth]{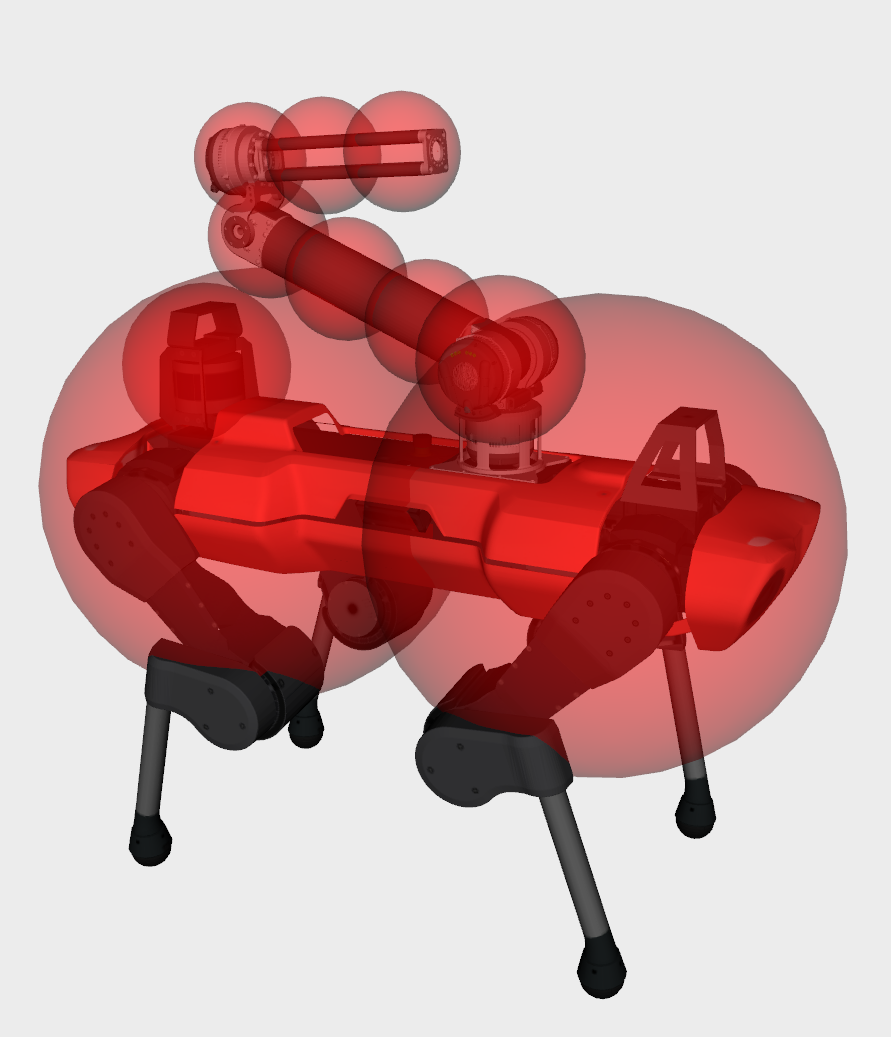}
  \end{minipage}
  \caption{Robot detailed (left) and simplified (middle) collision model. In the detailed model, the torso comprises the following three links: face front, shell, and face rear, with 11 collision bodies in total. In the simplified model, one box encloses the torso and one cylinder combines the elbow and forearm. (right) Sphere approximations for the simplified model.}
  \label{fig:robotCollisionModeling}
  \vspace{-6mm}
\end{figure}

We only consider potential collisions between the arm and the torso, since the arm rarely collides with the legs due to the mechanical joint limits. Therefore, no collision primitives are defined on robot's legs. Ideally, one would use collision bodies of various shapes and sizes to capture as many details as possible. In this way, more precise collision distances are available for collision avoidance. We describe the base in detail with thirteen collision bodies, consisting of eleven for the torso, one for the front handle, and one for the LiDAR cage as depicted in \figref{fig:robotCollisionModeling}. The collision model of the arm consists of three collision bodies, each for the upper arm, elbow, and forearm.

In comparison, simplified modeling is beneficial to easing computation. We trim off trivial details by replacing the eleven torso collision bodies with one big box and grouping the elbow and forearm with one cylinder. This simplified modeling requires only three and two collision bodies for the base and the arm, respectively. The middle figure in \figref{fig:robotCollisionModeling} visualizes the simplified collision model.
\newline
\subsubsection{Strategies}
\label{SelfCollisionAvoidanceStrategies}
Given the two query approaches and the two modeling fashions, we propose the following strategies:
\\
For detailed modeling, query the distances between
    \begin{enumerate}[label=\alph*.]
        \item 39 collision pairs using the na{\"i}ve approach.
        \item 3 arm collision bodies and 1 broad-phase manager for 13 base collision bodies.
    \end{enumerate}
For simplified modeling, query the distances between
    \begin{enumerate}[label=\alph*., resume]
        \item 6 collision pairs  using the na{\"i}ve approach.
        \item 2 arm collision bodies and 1 broad-phase manager for 3 base collision bodies.
    \end{enumerate}
In \secref{BenchmarkSelfCollision}, we examine how each modeling and query fashion affects the MPC performance.
\subsection{Environment-Collision Avoidance}
\label{EnvironmentCollisionAvoidance}
Similar to \cite{CHOMP2, Johannes, Magnus, Mayank}, we enclose the robot with a set of collision spheres and query the signed distances at the sphere centers in the SDF. Each sphere contributes one collision constraint \eqref{eq:IneqDistConstraint} with the distance function $d_i$ and threshold $\epsilon_i$ defined as:
\begin{subequations}
\begin{align}
    d_i(\bm x_r, t) & = SDF({}_\mathcal{I}\bm p_i(\bm x_r,t)) \\
    \epsilon_i & = r_i,
\end{align}
\end{subequations}
where ${}_\mathcal{I}\bm p_i$ denotes the sphere center, a function of the robot configuration, and $r_i$ is the corresponding sphere radius.
\newline
\subsubsection{Automatic Sphere Approximation}
\label{AutomaticSphereApproximation}
Unlike prior works manually approximating the robot's collision bodies into spheres, we implement an automatic sphere approximation algorithm \cite{AutomaticSphere} fitting a box or a cylinder with enclosing spheres of the same radius. The approximation respects a user-defined maximal distance $\delta_{max}$ by which a sphere's surface can exceed from the collision body surface. It allows us to reuse the collision primitives defined for self-collision avoidance, leading to straightforward generalizations of this collision-free motion planner over various robots. In this work, we avoid the obstacles by applying the sphere approximation algorithm to the simplified model in \secref{RobotCollisionModeling}. The visualization of the approximation result is shown in \figref{fig:robotCollisionModeling}.

\subsubsection{ESDF Mapping}
\label{EsdfMapping}

We use the \emph{FIESTA} mapping algorithm to create an SDF map of the scene~\cite{FIESTA}. Using point-cloud information and robot's odometry, the algorithm first creates an occupancy map and uses it to compute the SDF map. Similar to~\cite{Mayank}, we use tri-linear interpolation to query distances and compute gradients which are then cached to save computation time during MPC planning. We show in \secref{BenchmarkEnvironmentCollision} that this methodology adds little computation cost compared to the blind case. 

\section{SYSTEM DESCRIPTION}
The proposed MPC framework and the underlying whole-body controller (WBC) are robot agnostic. The rigid body dynamics of the robot is handled by the \emph{Pinocchio} C++ library \cite{pinocchio1, pinocchio2}. The distance queries between collision primitives are supported by the \emph{Flexible Collision Library} (\emph{FCL}) \cite{FCL} based on the Gilbert-Johnson-Keerthi (GJK) algorithm \cite{GJK}. The SLQ algorithm for the MPC planner is based on the \emph{OCS2} toolbox \cite{OCS2}, which implements efficient and numerically stable optimal control for switched systems. 

We perform several hardware tests with the ANYmal C platform equipped with DynaArm, a torque-controllable 4-DoF robotic arm. Together with powerful actuators, DynaArm is capable of highly dynamic manipulation with a payload capability of up to $7$~kg. The robot's onboard computer (Intel Core i7-8850H CPU@4GHz hexacore processor) takes care of both the MPC-based planner and the WBC. The main control loop along with the state estimator are executed at $400$ Hz. The MPC loop plans with a time horizon of $1$ s, at an update rate of $70$ Hz.
\section{RESULTS} \label{Results}
We perform various experiments to show the ability of the MPC formulation for legged mobile manipulation. In \secref{ExperimentsSelfCollision}, we show balancing with the arm and throwing a weight while respecting self-collision avoidance. In \secref{ExperimentsEnvironmentCollision}, we showcase environment-collision avoidance via SDF for tasks such as dynamic obstacle avoidance and door passing.

\subsection{Self-Collision Avoidance}
\label{ExperimentsSelfCollision}
In all the experiments, the RBF parameters are $\mu = 10^{-2}$ and $\delta = 10^{-3}$ for all collision pairs. We carry out several dynamic motions, which were only simulated in the previous work \cite{JeanPierre} due to the lack of self-collision avoidance.

\begin{figure}[t]
    \centering
    \begin{subfigure}[t]{0.5\textwidth}
        \centering
        \scalebox{0.75}{
        \begin{tikzpicture}
\node (A) at (0,0) {\includegraphics[trim={8cm 0cm 6cm 3cm},clip,scale=0.19]{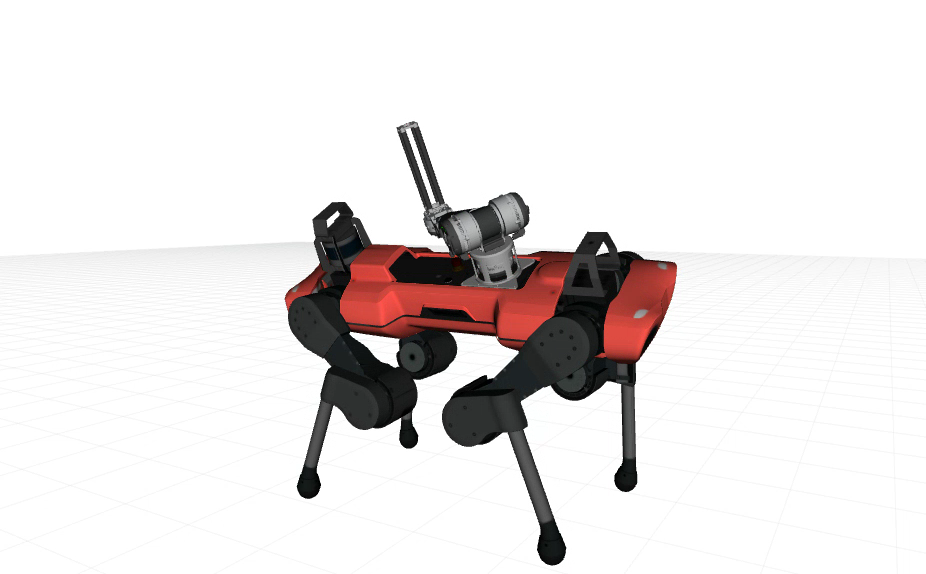}};
\node[right] (B) at (A.east) {\includegraphics[trim={8cm 0cm 6cm 3cm},clip,scale=0.19]{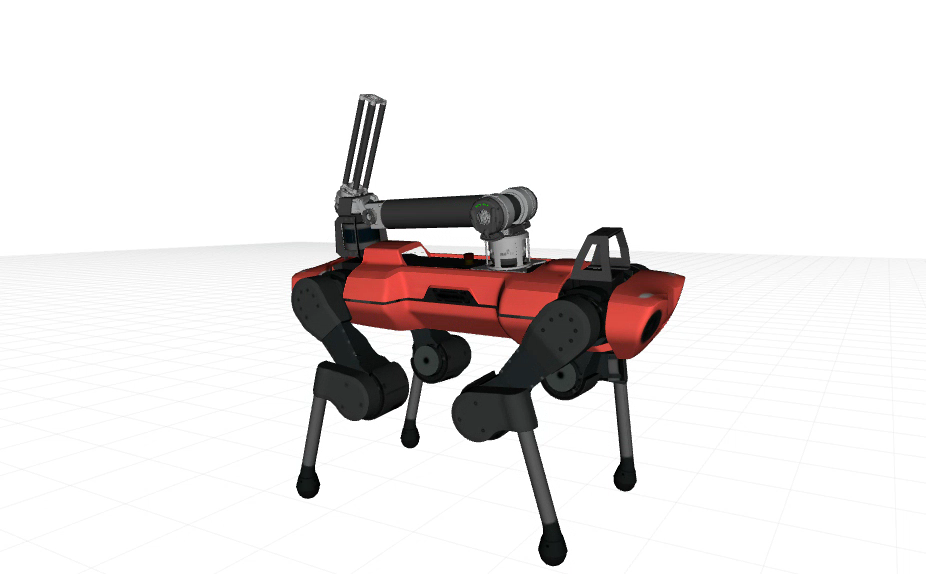}};
\node[right] (C) at (B.east) {\includegraphics[trim={8cm 0cm 6cm 3cm},clip,scale=0.19]{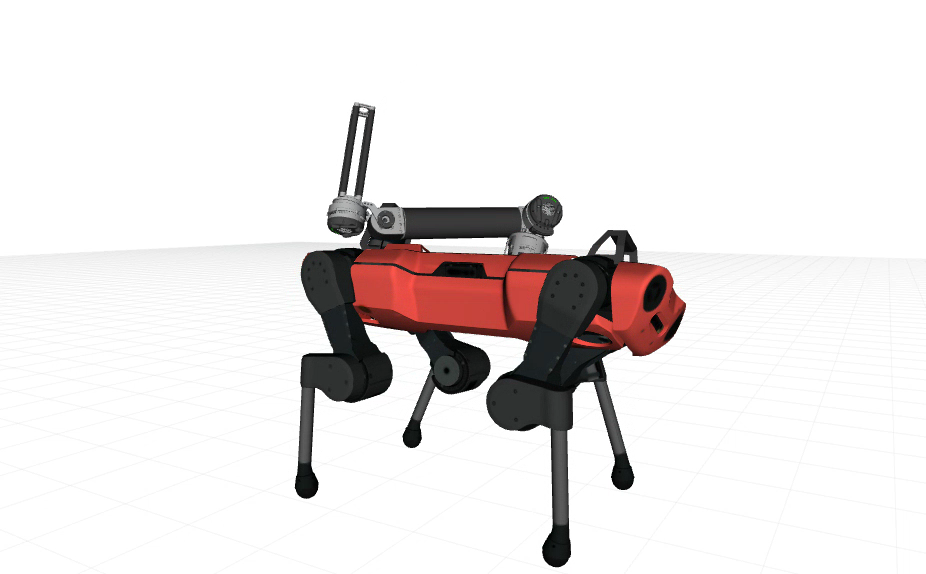}};
\node(AA) at (A.north west) {};
\node [ draw = black, minimum height = 0.3 cm, minimum width = 0.4cm, xshift = 0.675cm,yshift = -0.25cm, anchor = north east] at (AA) {\small 1};
\node(AA) at (B.north west) {};
\node [ draw = black, minimum height = 0.3 cm, minimum width = 0.4cm, xshift = 0.675cm,yshift = -0.25cm, anchor = north east] at (AA) {\small 2};
\node(AA) at (C.north west) {};
\node [ draw = black, minimum height = 0.3 cm, minimum width = 0.4cm, xshift = 0.675cm,yshift = -0.25cm, anchor = north east] at (AA) {\small 3};
\end{tikzpicture}}
        \label{fig:tail_sim}
    \end{subfigure}
    \vspace{-6mm}
    
    \begin{subfigure}[t]{0.5\textwidth}
        \centering
        \scalebox{0.75}{
      \begin{tikzpicture}
\node (A) at (0,0) {\includegraphics[trim={20cm 3cm 25cm 12cm},clip,scale=0.15]{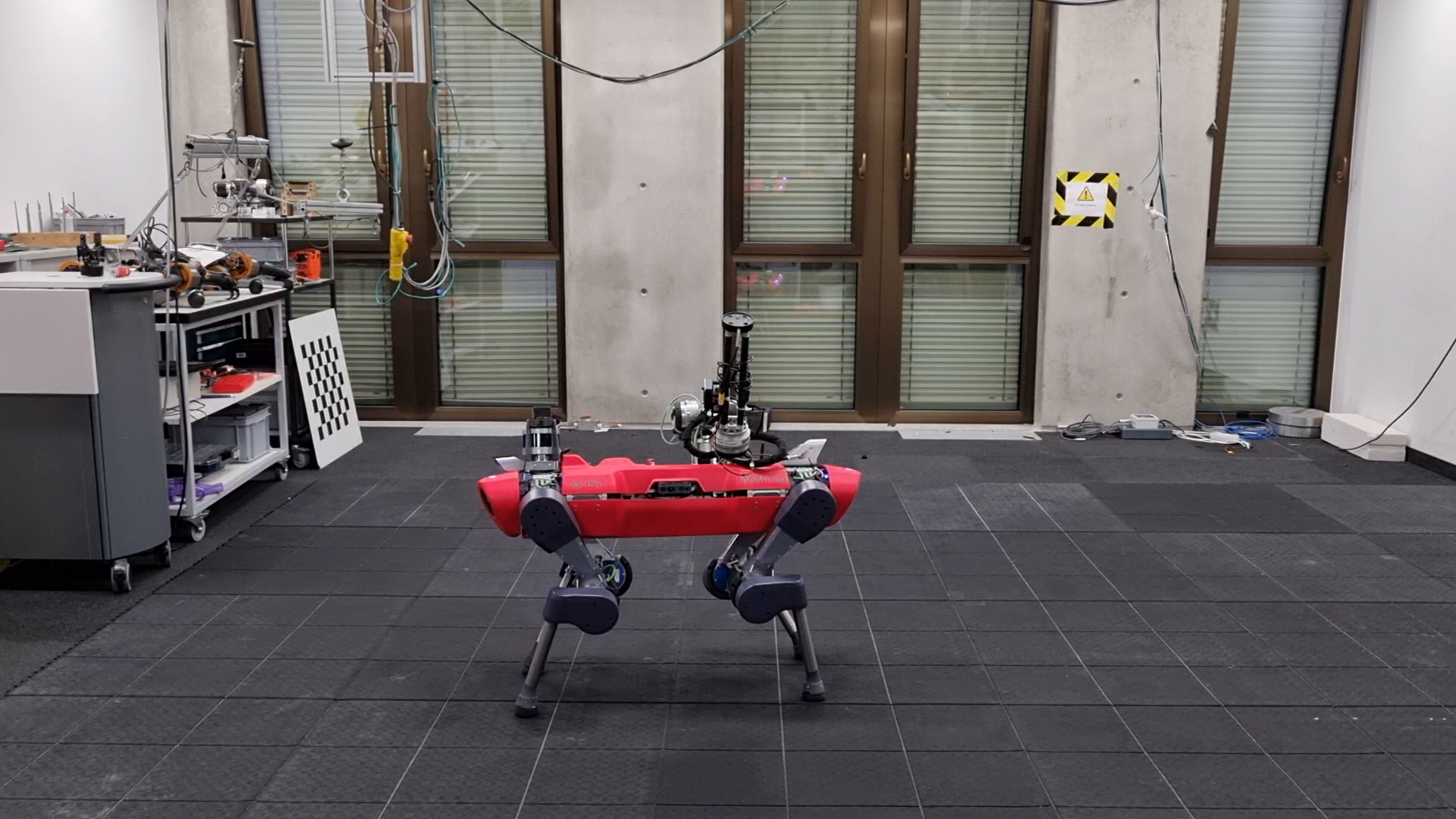}};
\node[right] (B) at (A.east) {\includegraphics[trim={20cm 3cm 25cm 12cm},clip,scale=0.15]{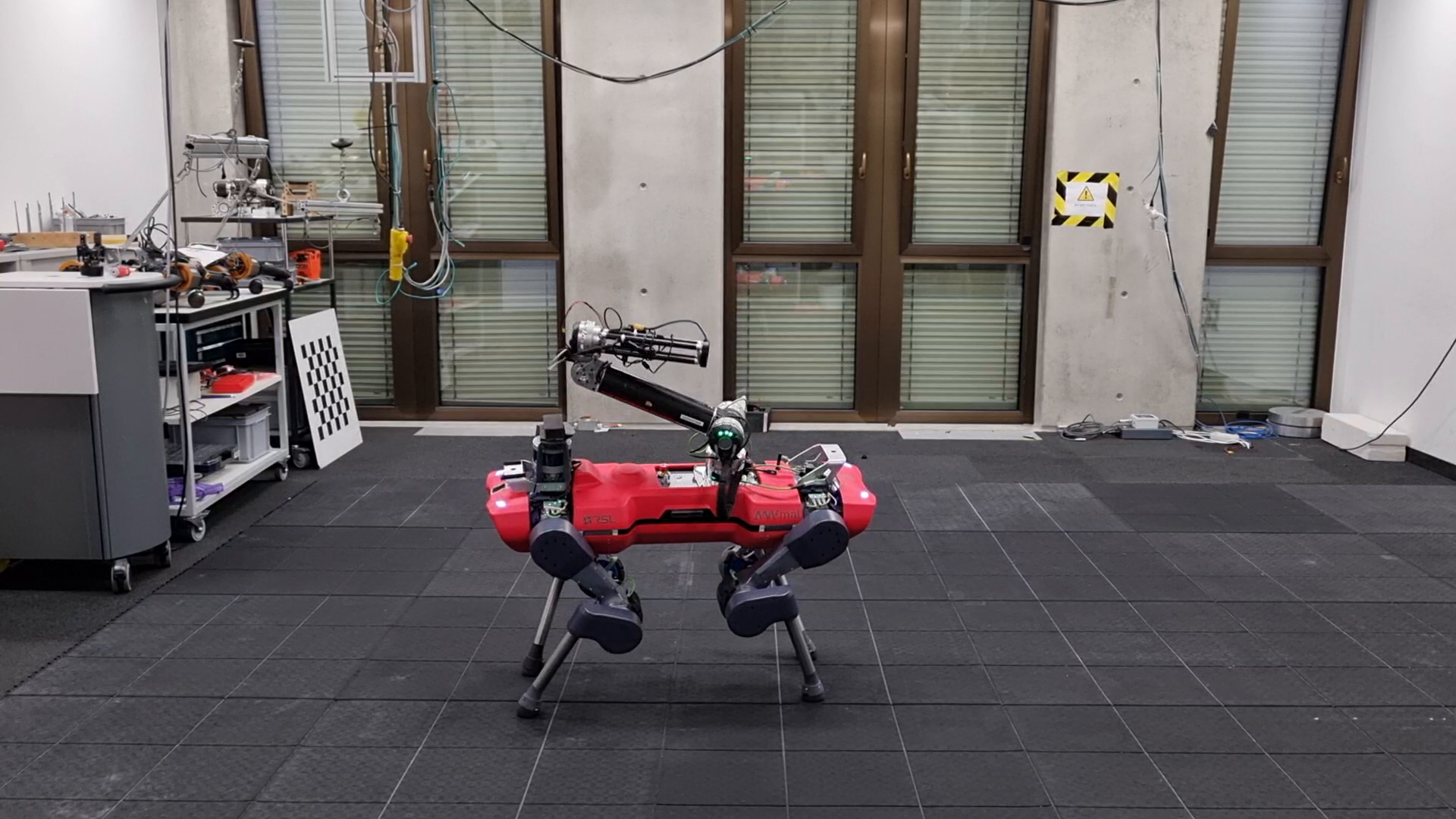}};
\node[right] (C) at (B.east) {\includegraphics[trim={20cm 3cm 25cm 12cm},clip,scale=0.15]{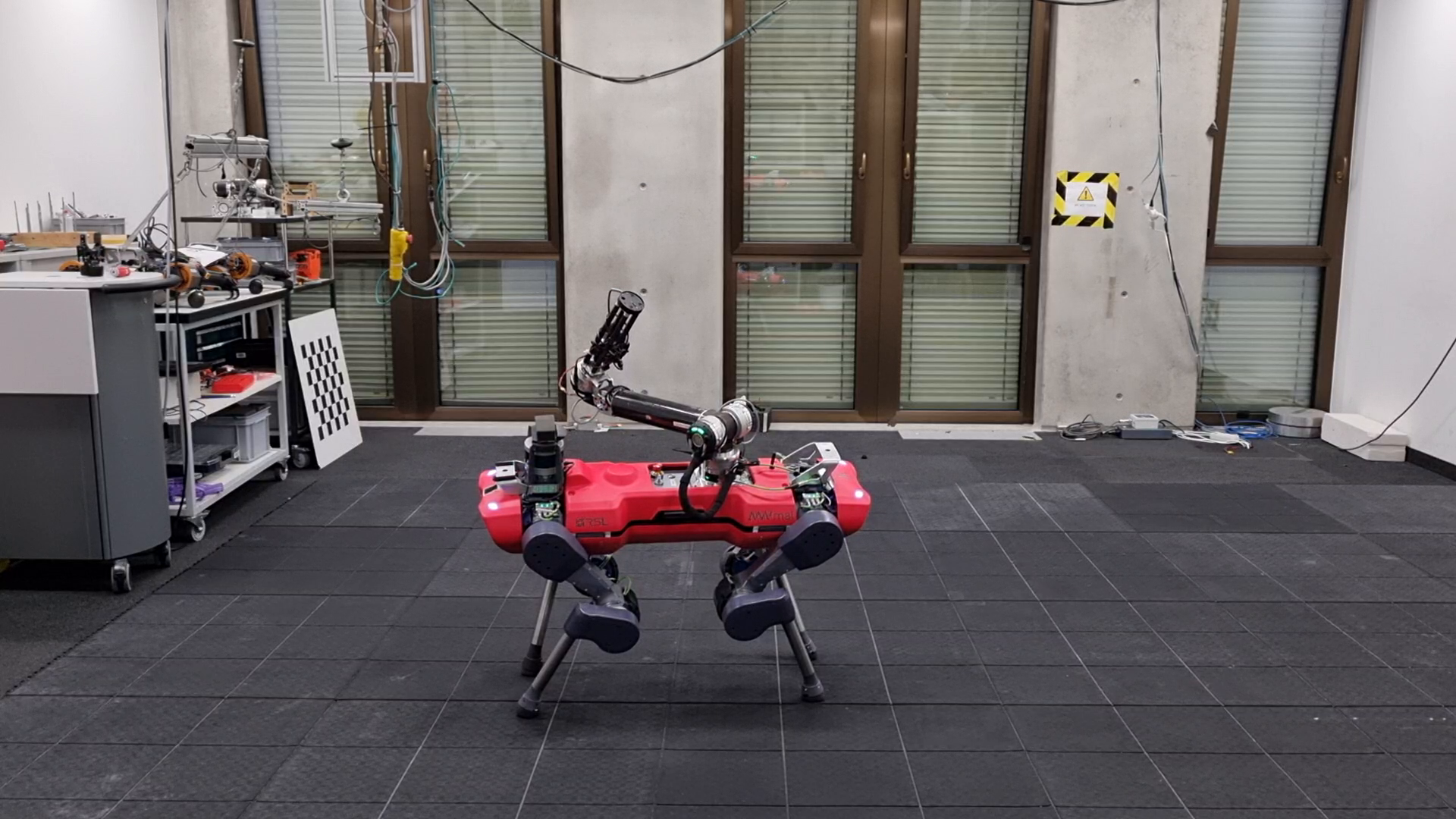}};
\node(AA) at (A.north west) {};
\node [ draw = white, minimum height = 0.3 cm, minimum width = 0.4cm, xshift = 0.675cm,yshift = -0.25cm, anchor = north east] at (AA) {\small\color{white} 4};
\node(AA) at (B.north west) {};
\node [ draw = white, minimum height = 0.3 cm, minimum width = 0.4cm, xshift = 0.675cm,yshift = -0.25cm, anchor = north east] at (AA) {\small\color{white} 5};
\node(AA) at (C.north west) {};
\node [ draw = white, minimum height = 0.3 cm, minimum width = 0.4cm, xshift = 0.675cm,yshift = -0.25cm, anchor = north east] at (AA) {\small\color{white} 6};
\end{tikzpicture}}
        \label{fig:tail_exp_roll}
    \end{subfigure}
    \caption{(1)-(3) Blind simulated and (4)-(6) collision-free real-world balancing with the arm when the base roll angle changes between $\pm20^{o}$. Self-collision may occur to the blind robot between its arm and LiDAR in (2), whereas the arm swiftly avoids collision in (5).}
    \label{fig:tail_balancing}
\end{figure}
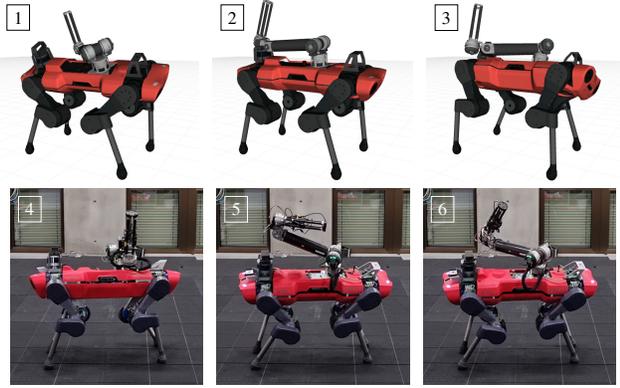

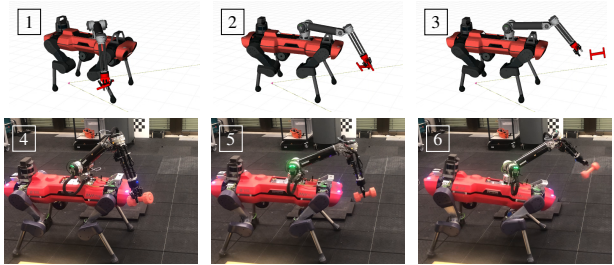
\begin{figure}[t]
    \centering
    \begin{subfigure}[t]{0.5\textwidth}
        \centering
        \scalebox{0.75}{
        \begin{tikzpicture}
\node (A) at (0,0) {\includegraphics[trim={2cm 4cm 13cm 6cm},clip,scale=0.19]{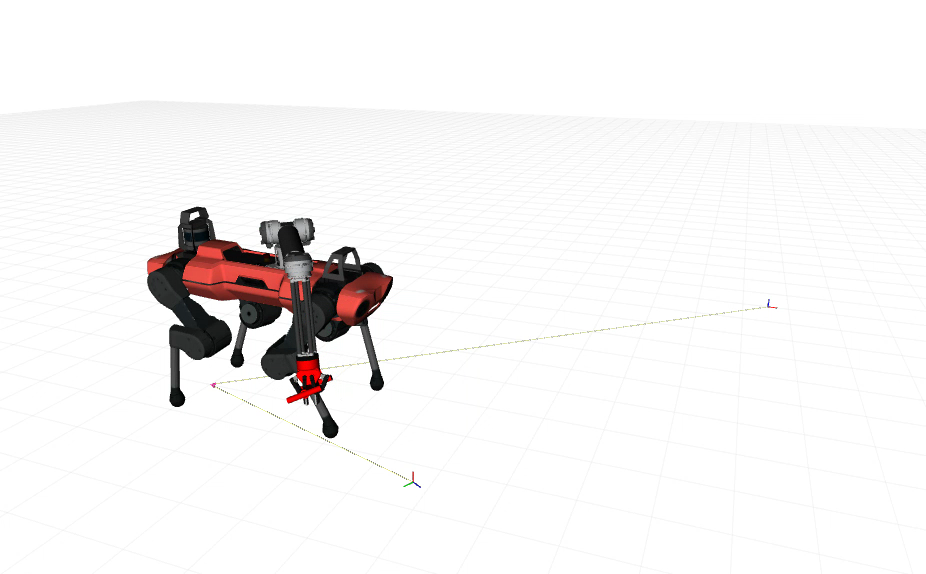}};
\node[right] (B) at (A.east) {\includegraphics[trim={2cm 4cm 13cm 6cm},clip,scale=0.19]{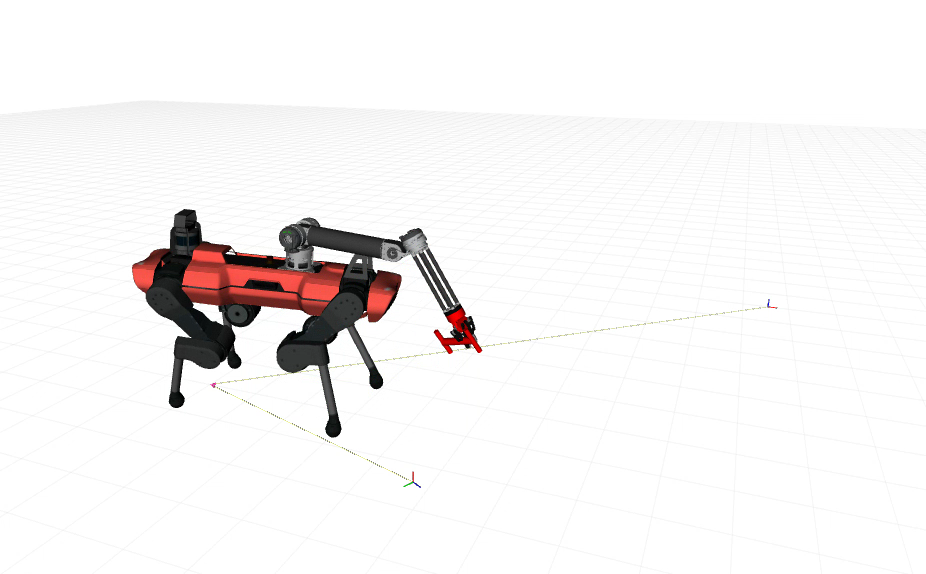}};
\node[right] (C) at (B.east) {\includegraphics[trim={2cm 4cm 13cm 6cm},clip,scale=0.19]{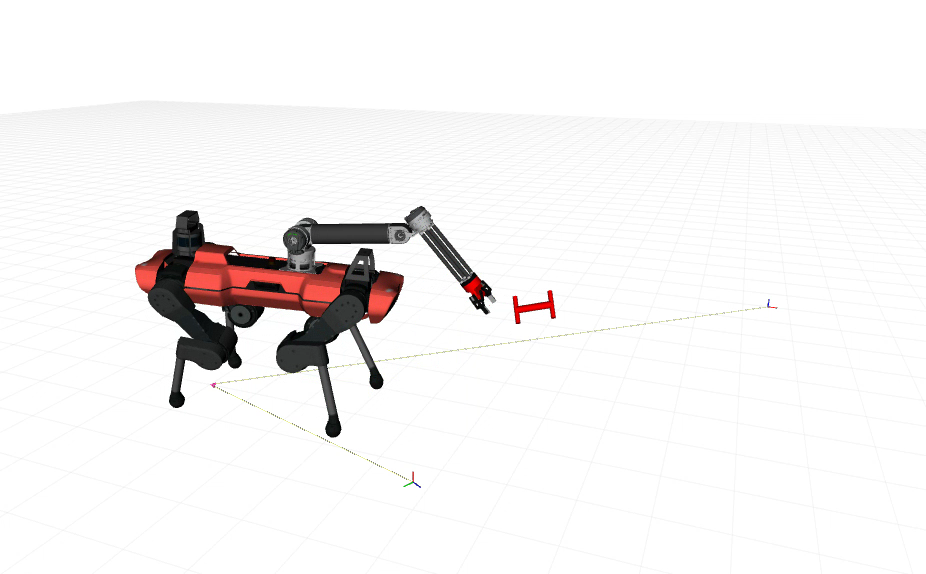}};
\node(AA) at (A.north west) {};
\node [ draw = black, minimum height = 0.3 cm, minimum width = 0.4cm, xshift = 0.675cm,yshift = -0.25cm, anchor = north east] at (AA) {\small 1};
\node(AA) at (B.north west) {};
\node [ draw = black, minimum height = 0.3 cm, minimum width = 0.4cm, xshift = 0.675cm,yshift = -0.25cm, anchor = north east] at (AA) {\small 2};
\node(AA) at (C.north west) {};
\node [ draw = black, minimum height = 0.3 cm, minimum width = 0.4cm, xshift = 0.675cm,yshift = -0.25cm, anchor = north east] at (AA) {\small 3};
\end{tikzpicture}}
        \label{fig:throwing_sim_front}
    \end{subfigure}
    \vspace{-6mm}
    
    \begin{subfigure}[t]{0.5\textwidth}
        \centering
        \scalebox{0.75}{
      \begin{tikzpicture}
\node (A) at (0,0) {\includegraphics[trim={0cm 0cm 25cm 6cm},clip,scale=0.08]{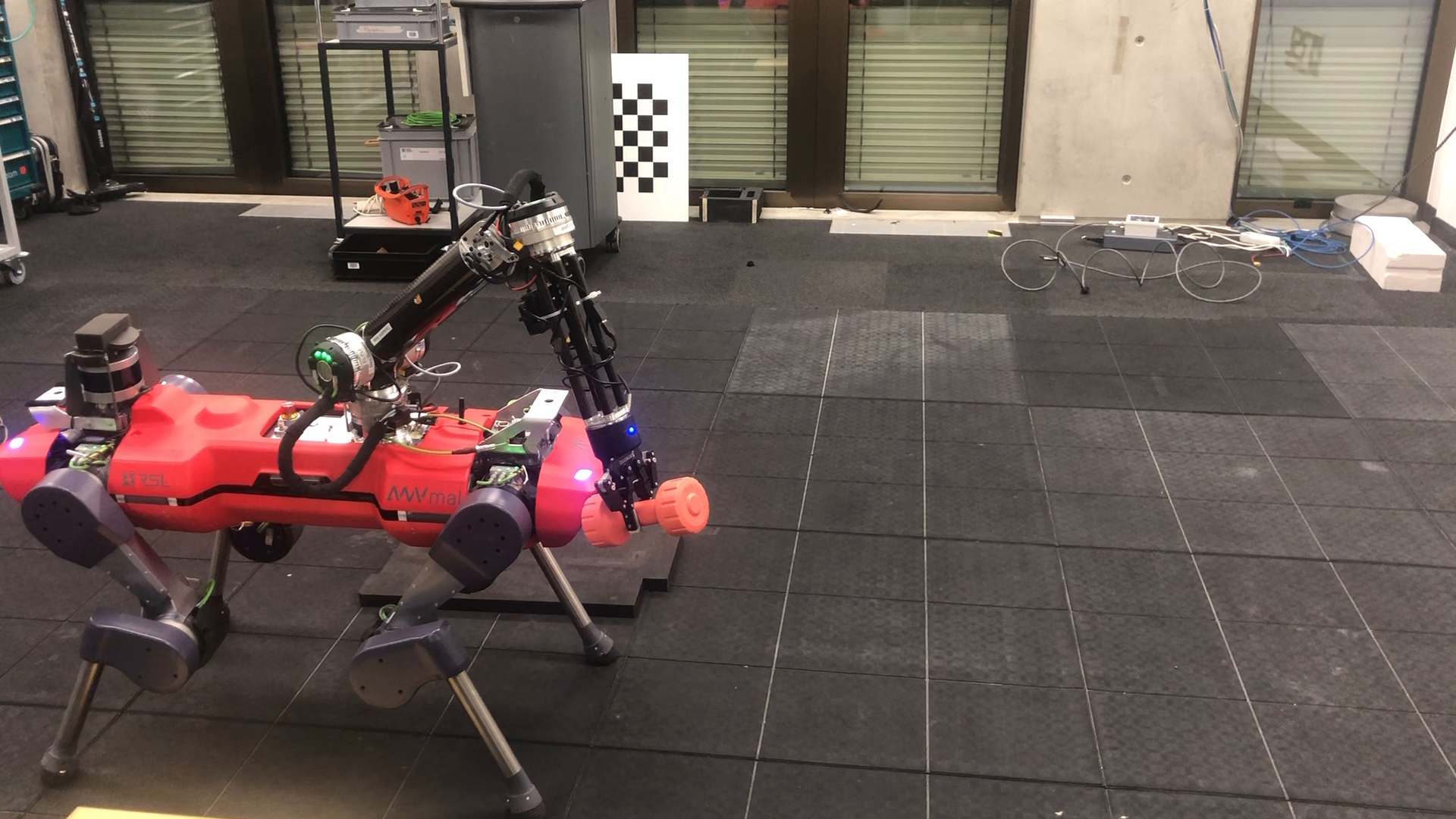}};
\node[right] (B) at (A.east) {\includegraphics[trim={0cm 0cm 25cm 6cm},clip,scale=0.08]{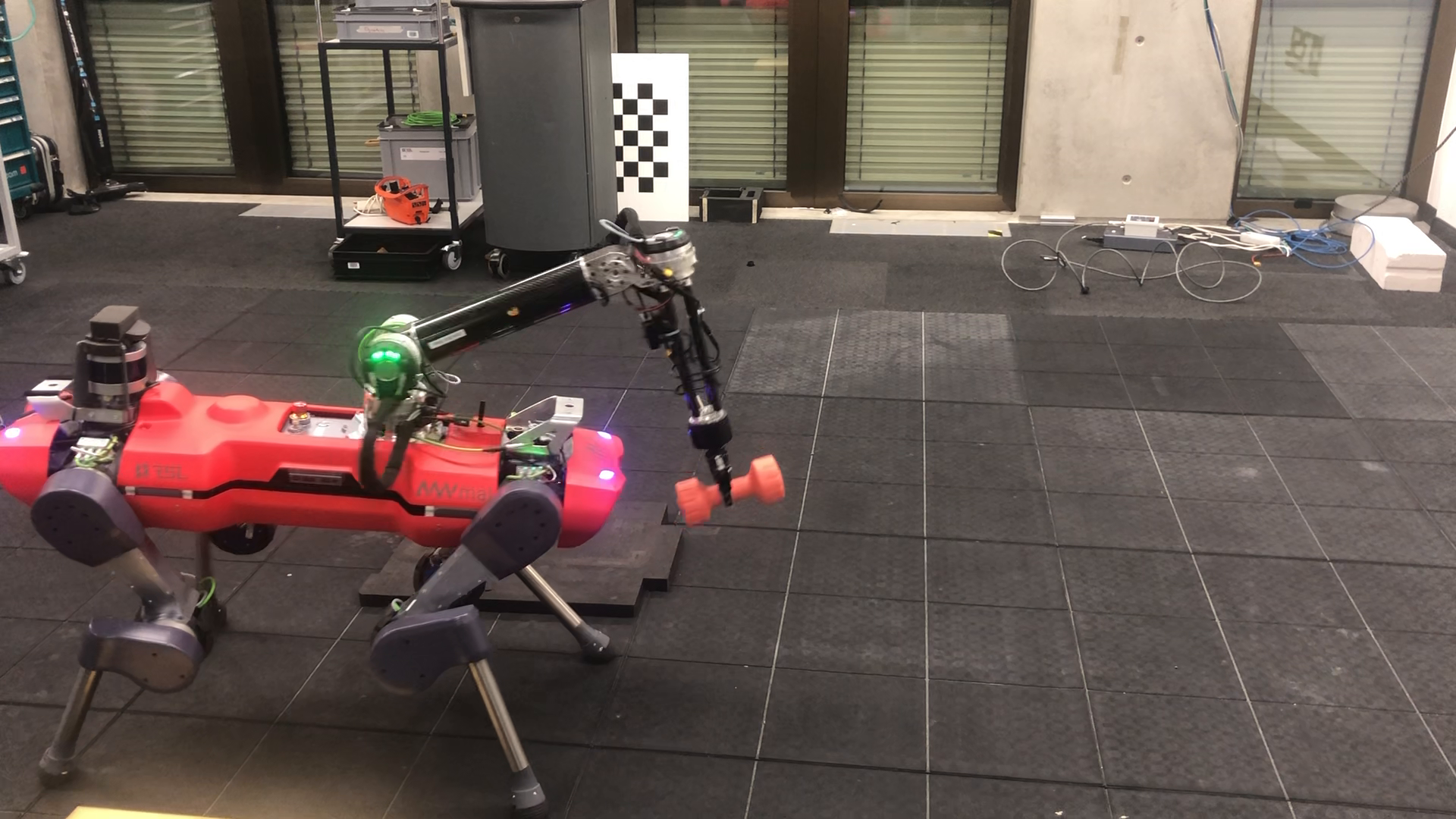}};
\node[right] (C) at (B.east) {\includegraphics[trim={0cm 0cm 25cm 6cm},clip,scale=0.08]{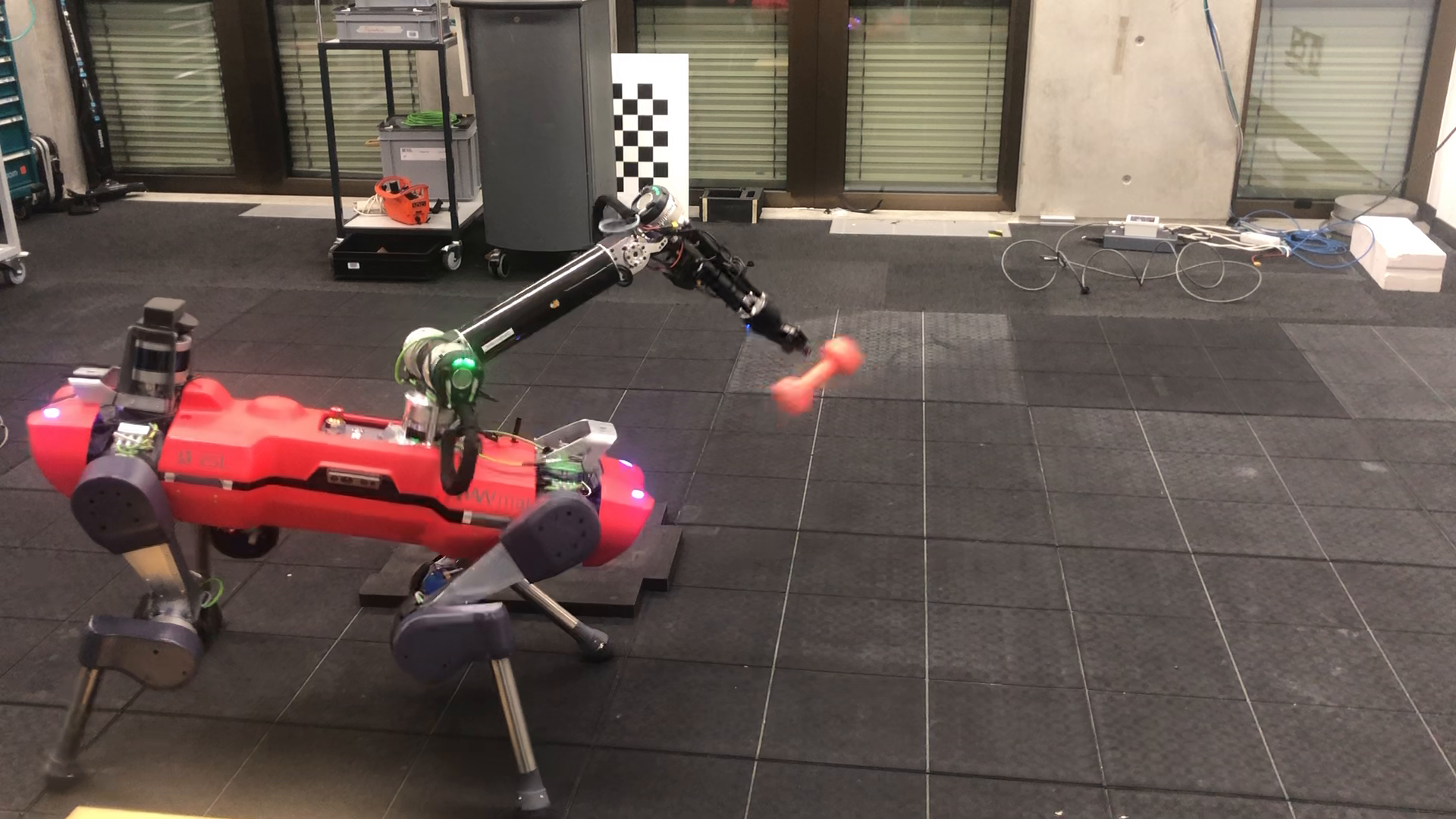}};
\node(AA) at (A.north west) {};
\node [ draw = white, minimum height = 0.3 cm, minimum width = 0.4cm, xshift = 0.675cm,yshift = -0.25cm, anchor = north east] at (AA) {\small\color{white} 4};
\node(AA) at (B.north west) {};
\node [ draw = white, minimum height = 0.3 cm, minimum width = 0.4cm, xshift = 0.675cm,yshift = -0.25cm, anchor = north east] at (AA) {\small\color{white} 5};
\node(AA) at (C.north west) {};
\node [ draw = white, minimum height = 0.3 cm, minimum width = 0.4cm, xshift = 0.675cm,yshift = -0.25cm, anchor = north east] at (AA) {\small\color{white} 6};
\end{tikzpicture}}
        \label{fig:throwing_exp_front}
    \end{subfigure}
    \caption{(1)-(3) Simulation of blind forward weight throwing while in stance with potential collision between the arm and handle in (2). (4)-(6) Hardware collision-free demonstration.}
    \label{fig:throwing}
    \vspace{-6mm}
\end{figure}

\paragraph*{Balancing with Arm} For free-motion modes, we use the manipulator as a tail to balance dynamic base motions with $\alpha_2 = 1$ and $\alpha_1 = \alpha_3 = 0$. By reducing the penalties on the arm joint positions and velocities, we grant the MPC more freedom to exploit the wide range of motion and the high speed of the Dynaarm for balancing. For the first example, we command the robot to instantly change the base roll angle from $+20^{o}$ to $-20^{o}$ while standing. As shown in \figref{fig:tail_balancing}, quick changes in the base orientation may lead to a collision between the arm and the LiDAR in the simulated blind robot. On the contrary, with self-collision avoidance, the manipulator does not only maintain its high speed but rapidly avoids crashing into the LiDAR on the hardware. In the second scenario, the robot trots sideways at a relatively high speed and instantly switches directions. The arm also quickly steers away from the LiDAR while swinging in the opposite direction \figref{fig:teaser}. Both scenarios demonstrate the efficacy of self-collision avoidance in static and dynamic maneuvers.

\paragraph*{Weight Throwing} To fully exploit the capabilities of the MPC planner, we also showcase self-collision avoidance during object manipulation with $\alpha_3 = 1$ and $\alpha_1 = \alpha_2 = 0$. The robot is commanded to throw a $1.5$ kg dumbbell to the target position. The MPC planner optimizes the throwing motions according to the given switching time. In the first mission, the target position is $2$ m to the front and $2$ m to the left of the base center. As shown in \figref{fig:throwing}, the upper arm of the blind robot may collide with the front handle of the base in the simulation. However, with self-collision avoidance, the manipulator swiftly negotiates with the handle and prevents potential damage to the hardware. Finally, in the second scenario, the robot is tasked with throwing $2$ m to the back and $1$ m to the right while trotting, as shown in \figref{fig:teaser}. In this case, the throwing maneuver can be further enhanced by having the robot trot during the task execution, thus generating more dynamic movements through a wider range of base motions.

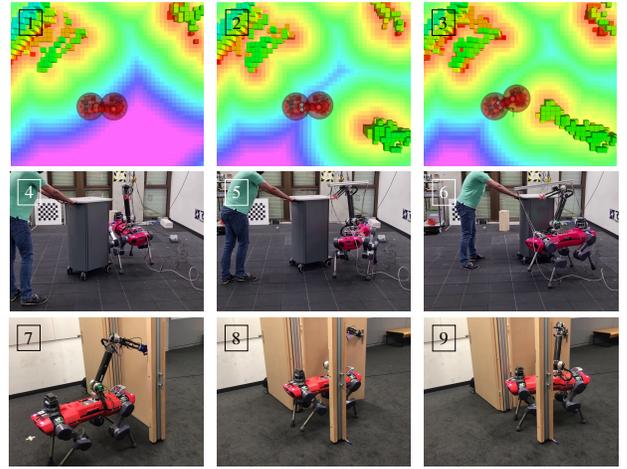
\begin{figure}[t]
    \centering
    \begin{subfigure}[t]{0.5\textwidth}
        \centering
        \scalebox{0.75}{
        \begin{tikzpicture}
\node (A) at (0,0) {\includegraphics[scale=0.08]{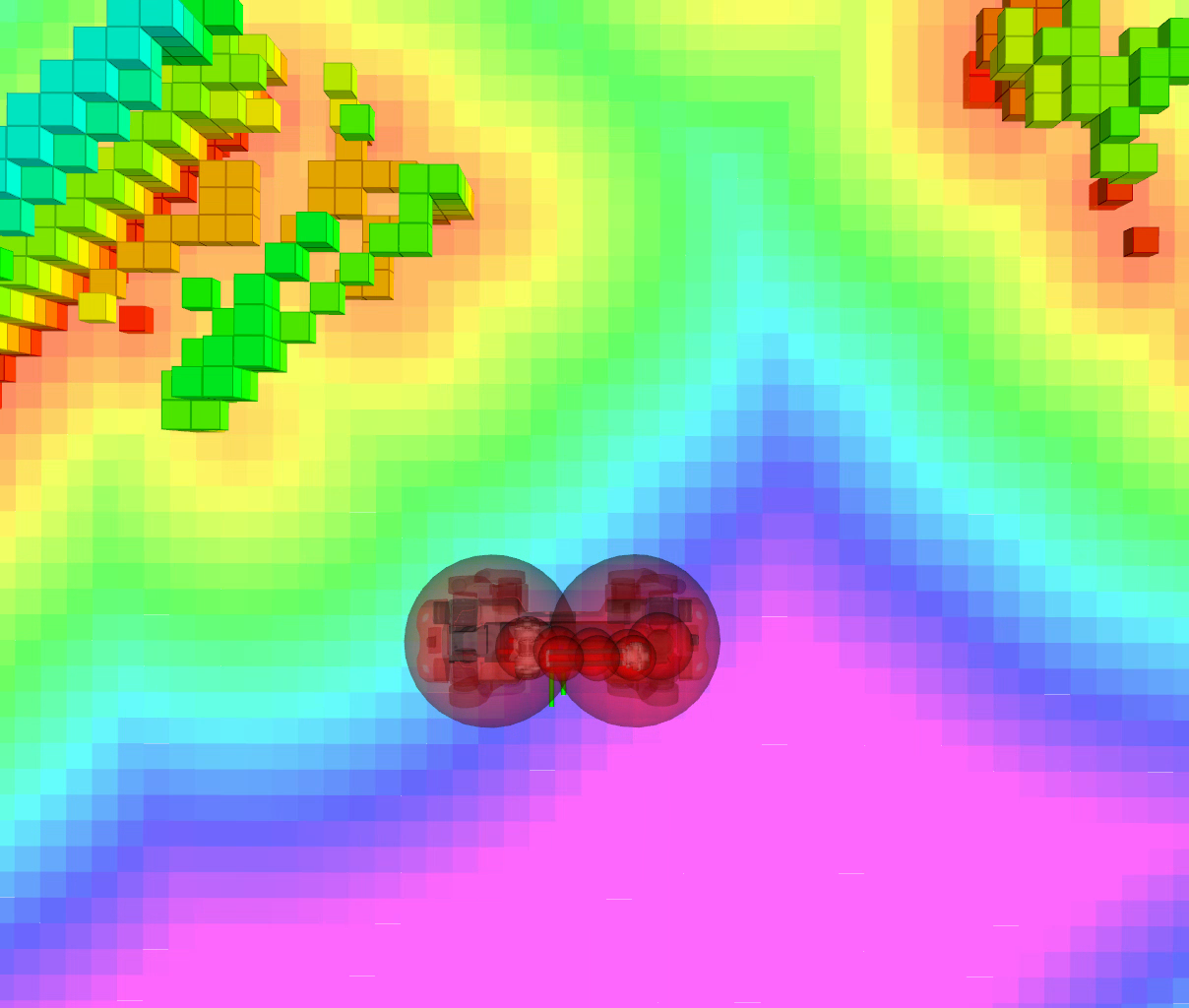}};
\node[right] (B) at (A.east) {\includegraphics[scale=0.08]{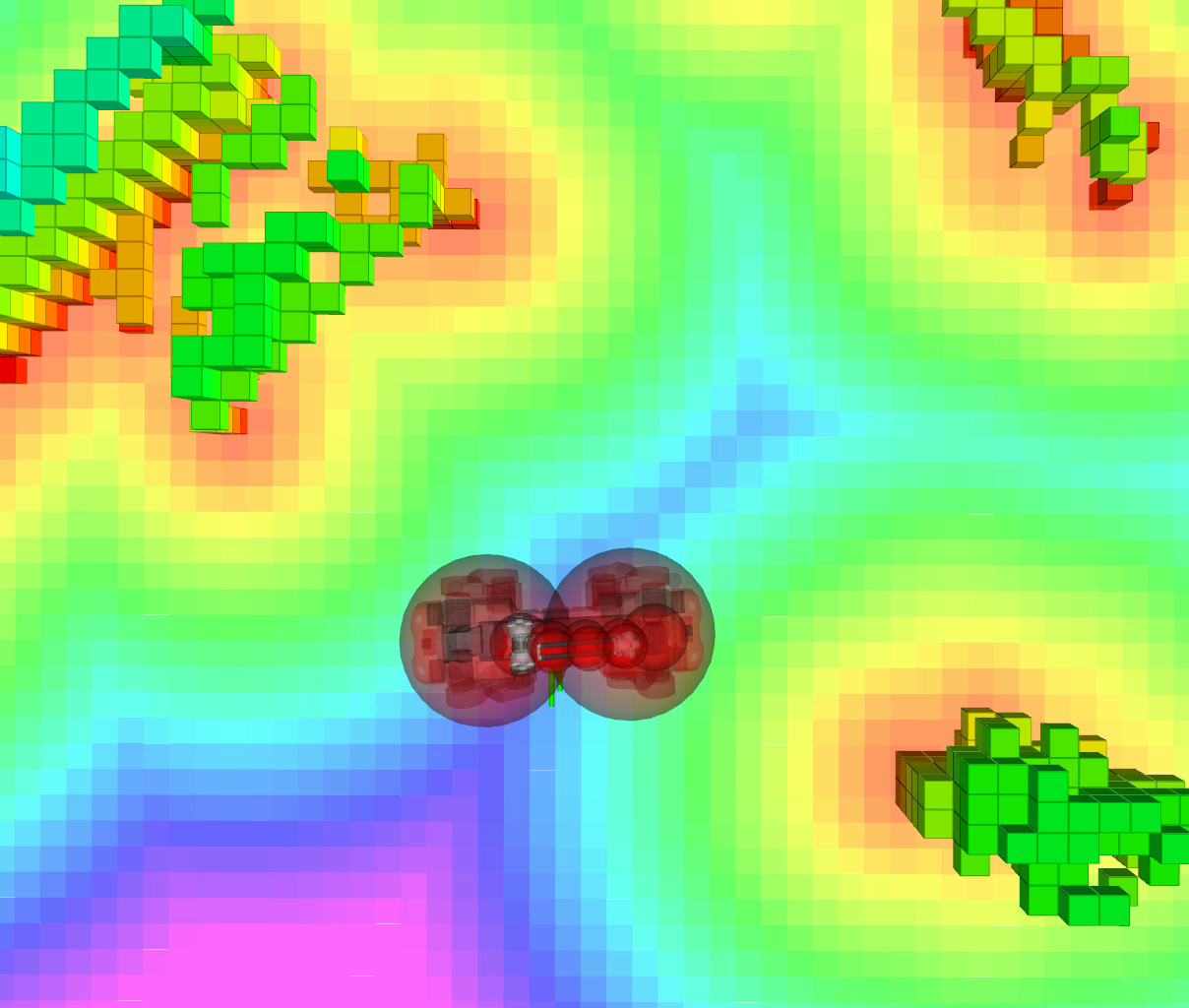}};
\node[right] (C) at (B.east) {\includegraphics[scale=0.08]{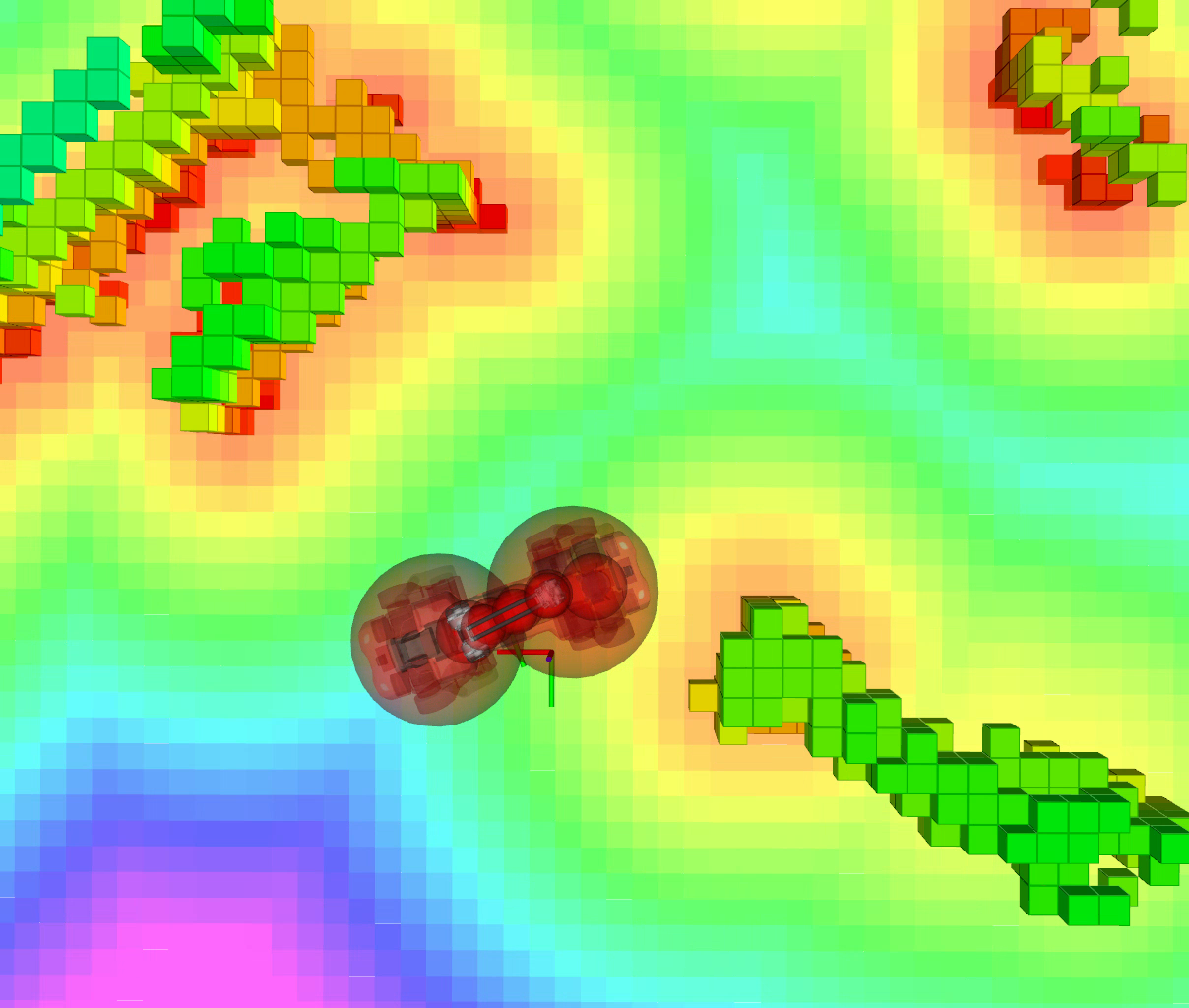}};
\node(AA) at (A.north west) {};
\node [ draw = black, minimum height = 0.3 cm, minimum width = 0.4cm, xshift = 0.675cm,yshift = -0.25cm, anchor = north east] at (AA) {\small 1};
\node(AA) at (B.north west) {};
\node [ draw = black, minimum height = 0.3 cm, minimum width = 0.4cm, xshift = 0.675cm,yshift = -0.25cm, anchor = north east] at (AA) {\small 2};
\node(AA) at (C.north west) {};
\node [ draw = black, minimum height = 0.3 cm, minimum width = 0.4cm, xshift = 0.675cm,yshift = -0.25cm, anchor = north east] at (AA) {\small 3};
\end{tikzpicture}}
        \label{fig:fiesta}
    \end{subfigure}
    \vspace{-6mm}
    
    \begin{subfigure}[t]{0.5\textwidth}
        \centering
        \scalebox{0.75}{
        \begin{tikzpicture}
\node (A) at (0,0) {\includegraphics[trim={8cm 0cm 6cm 3cm},clip,scale=0.11]{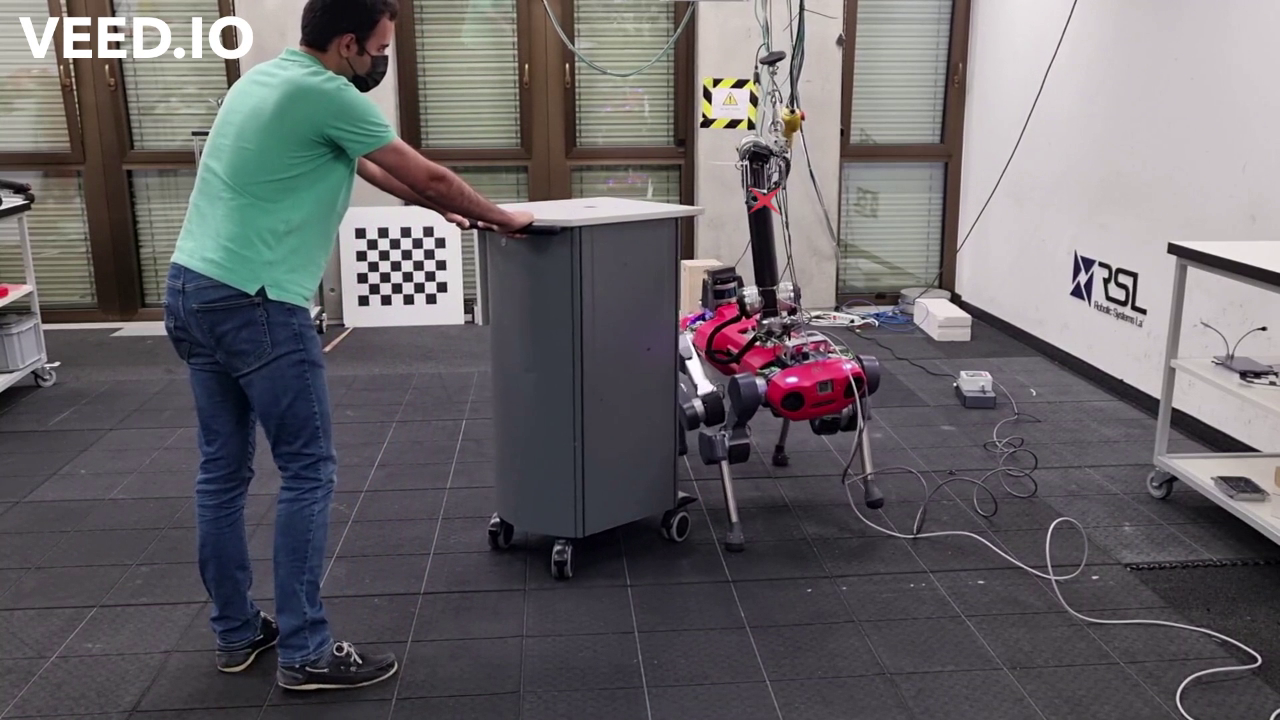}};
\node[right] (B) at (A.east) {\includegraphics[trim={8cm 0cm 6cm 3cm},clip,scale=0.11]{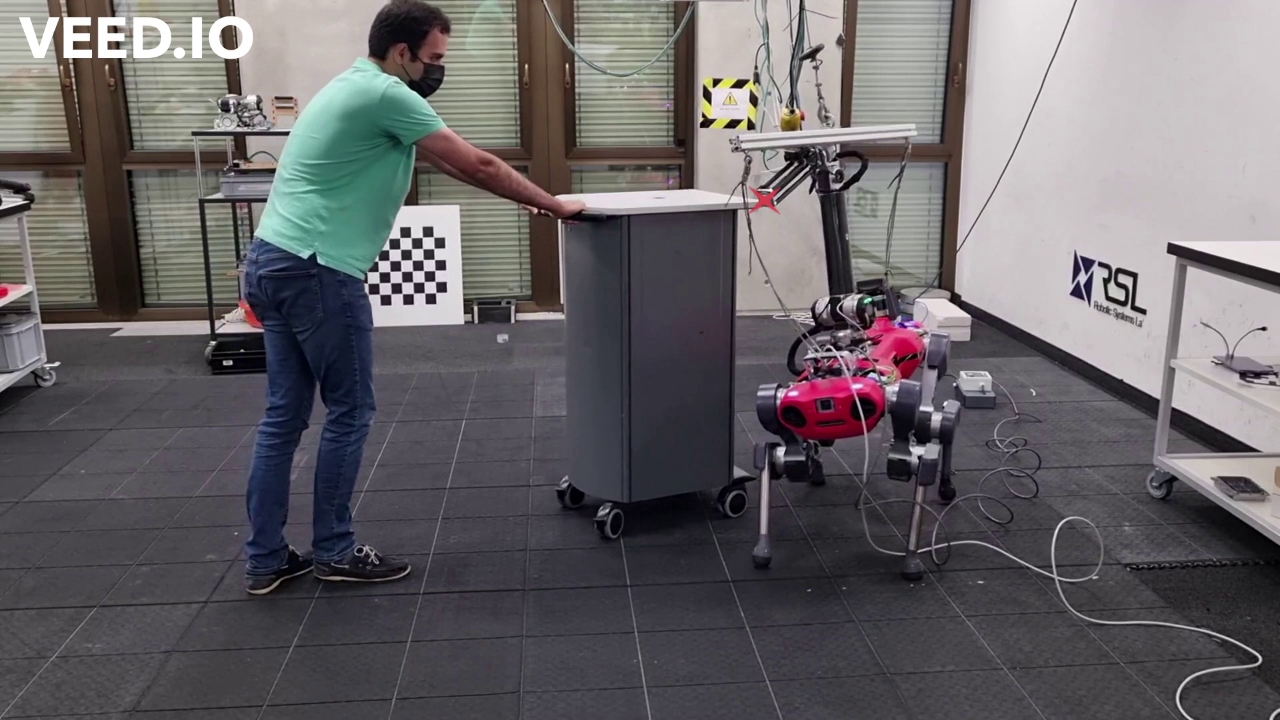}};
\node[right] (C) at (B.east) {\includegraphics[trim={8cm 0cm 6cm 3cm},clip,scale=0.11]{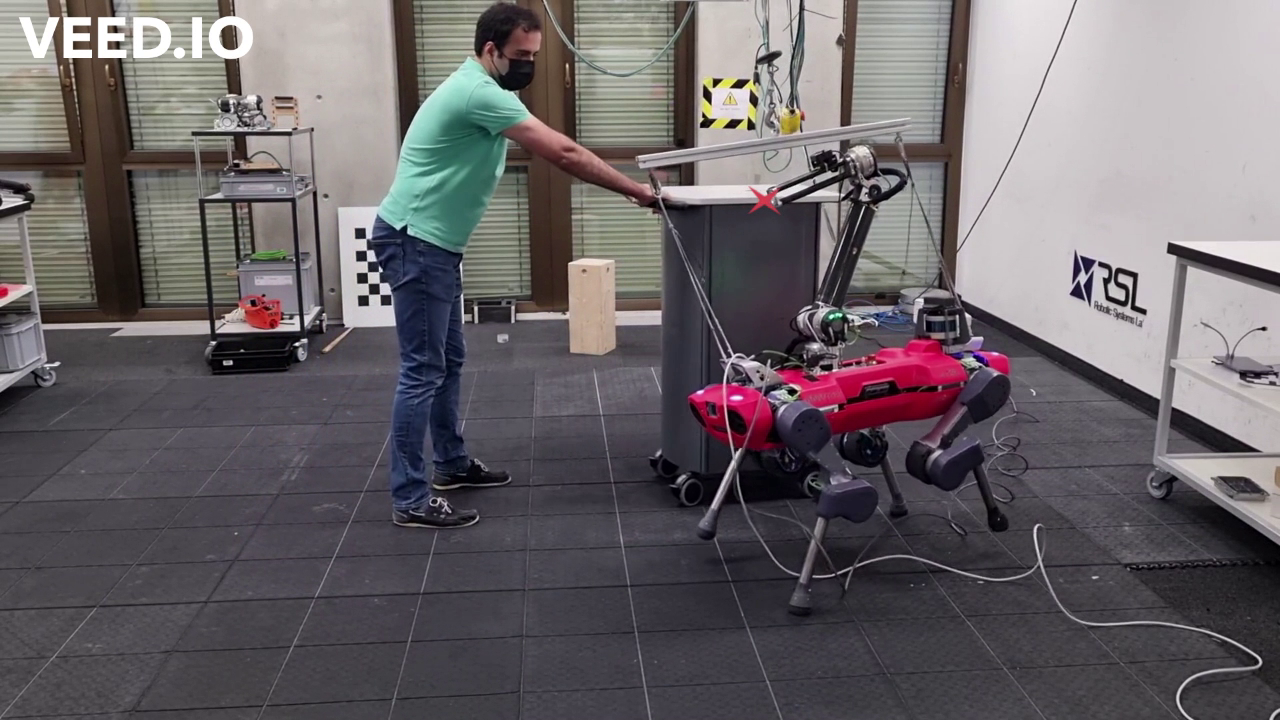}};
\node(AA) at (A.north west) {};
\node [ draw = white, minimum height = 0.3 cm, minimum width = 0.4cm, xshift = 0.675cm,yshift = -0.25cm, anchor = north east] at (AA) {\small\color{white} 4};
\node(AA) at (B.north west) {};
\node [ draw = white, minimum height = 0.3 cm, minimum width = 0.4cm, xshift = 0.675cm,yshift = -0.25cm, anchor = north east] at (AA) {\small\color{white} 5};
\node(AA) at (C.north west) {};
\node [ draw = white, minimum height = 0.3 cm, minimum width = 0.4cm, xshift = 0.675cm,yshift = -0.25cm, anchor = north east] at (AA) {\small\color{white} 6};
\end{tikzpicture}}
        \label{fig:moving_cart}
    \end{subfigure}
    \vspace{-6mm}
    
    \begin{subfigure}[t]{0.5\textwidth}
        \centering
        \scalebox{0.75}{
       \begin{tikzpicture}
\node (A) at (0,0) {\includegraphics[trim={3cm 0cm 15cm 0cm},clip,scale=0.069]{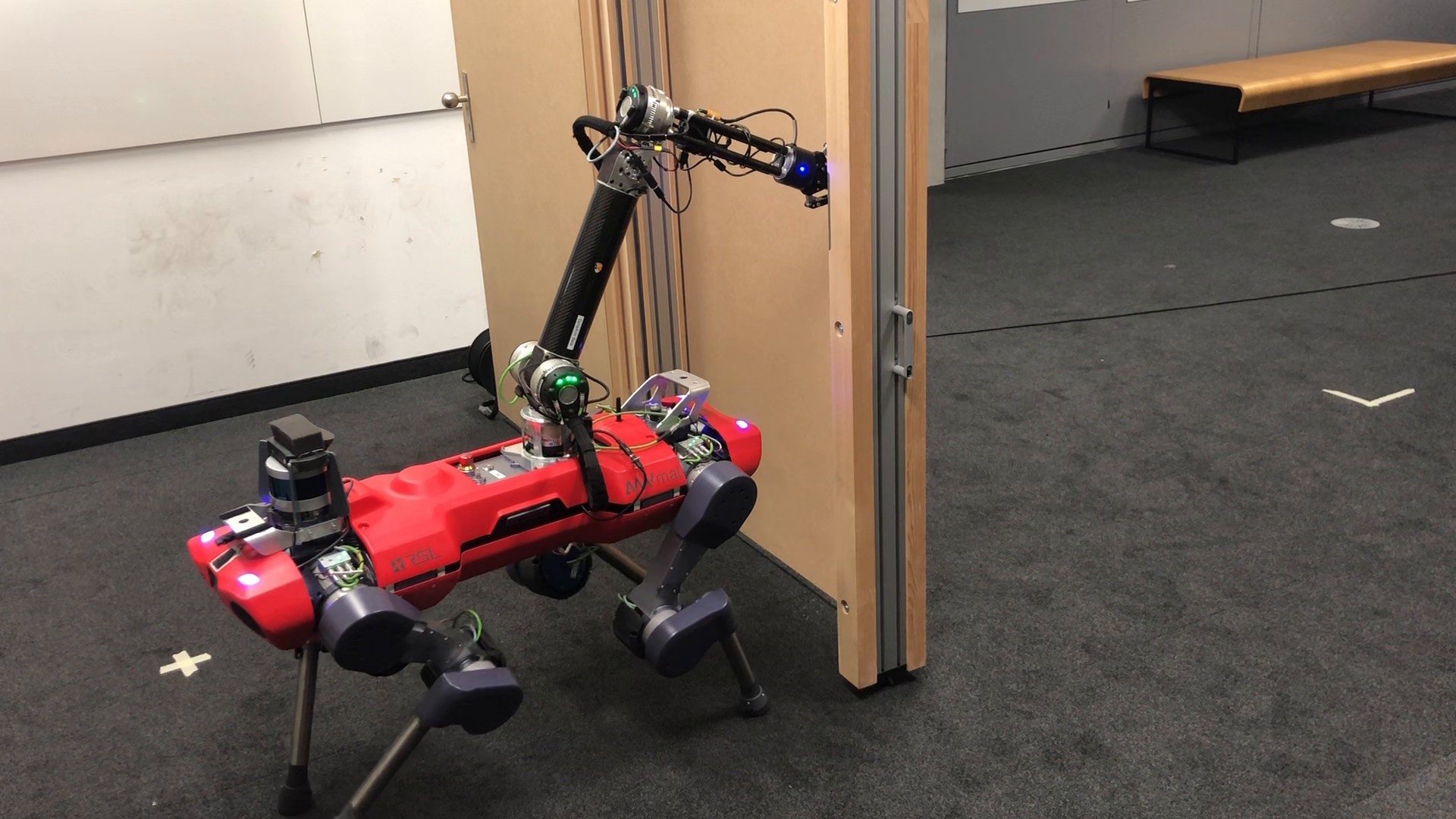}};
\node[right] (B) at (A.east) {\includegraphics[trim={10cm 0cm 8cm 0cm},clip,scale=0.069]{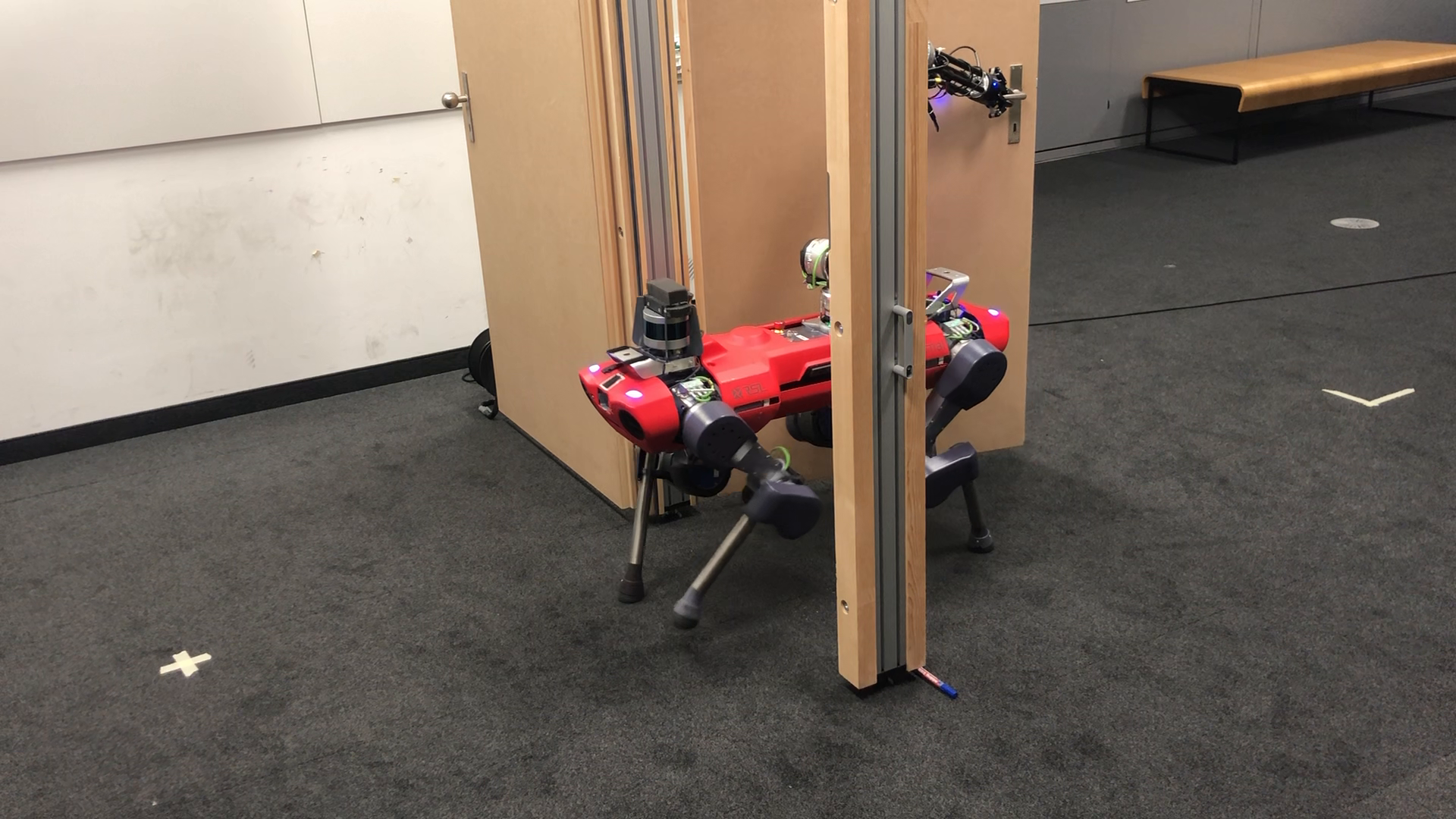}};
\node[right] (C) at (B.east) {\includegraphics[trim={10cm 0cm 8cm 0cm},clip,scale=0.069]{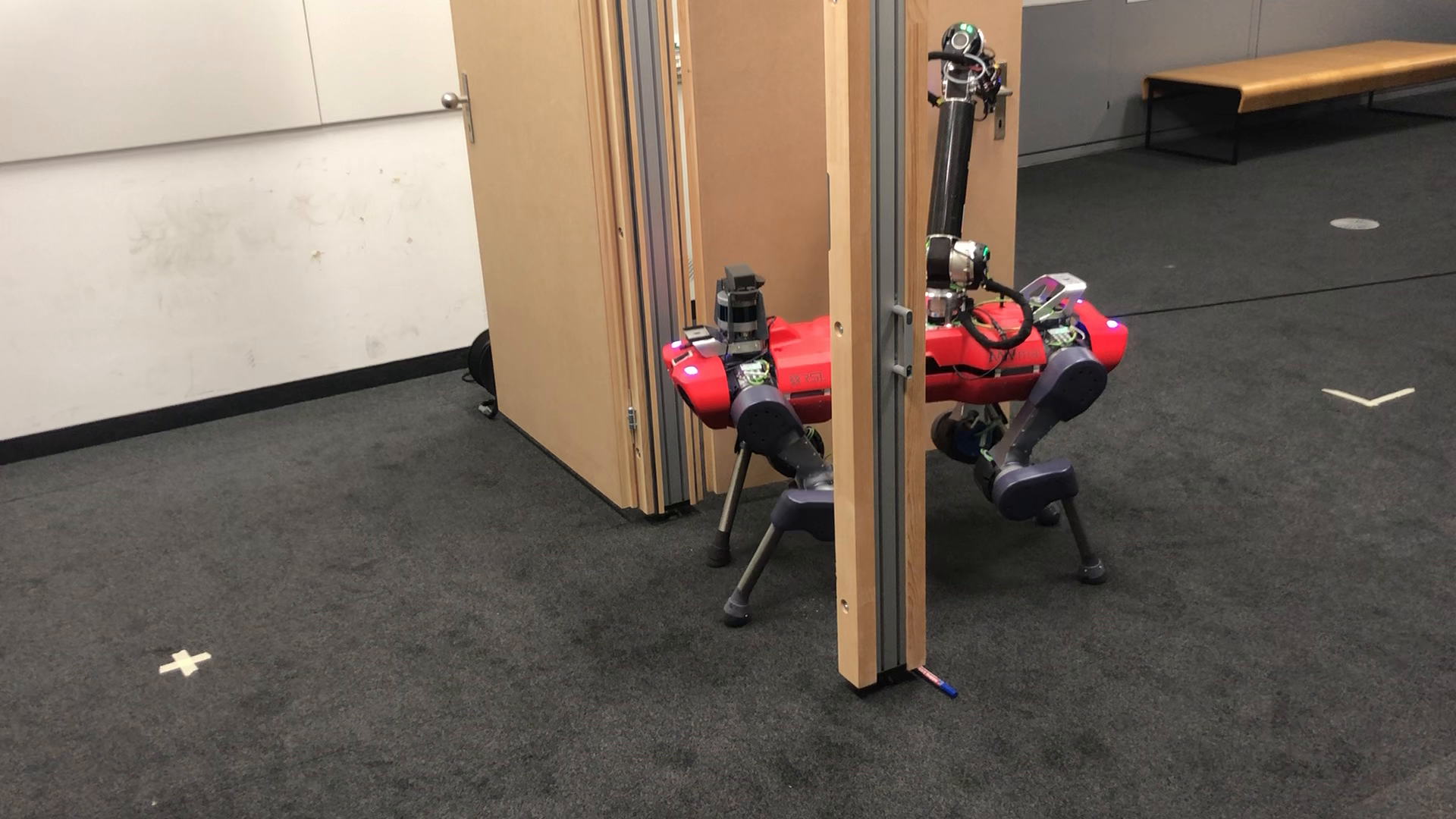}};
\node(AA) at (A.north west) {};
\node [ draw = black, minimum height = 0.3 cm, minimum width = 0.4cm, xshift = 0.675cm,yshift = -0.25cm, anchor = north east] at (AA) {\small 7};
\node(AA) at (B.north west) {};
\node [ draw = black, minimum height = 0.3 cm, minimum width = 0.4cm, xshift = 0.675cm,yshift = -0.25cm, anchor = north east] at (AA) {\small 8};
\node(AA) at (C.north west) {};
\node [ draw = black, minimum height = 0.3 cm, minimum width = 0.4cm, xshift = 0.675cm,yshift = -0.25cm, anchor = north east] at (AA) {\small 9};
\end{tikzpicture}}
        \label{fig:door_opening}
    \end{subfigure}
    \caption{(1)-(3) Visualization of the sphere approximation, ESDF slice, and occupancy grip map while avoiding an incoming obstacle. (4)-(6) Dynamic obstacle avoidance with an approaching cart while keeping track of an end-effector target position. (7)-(9) Autonomous door opening with collision-avoiding base yaw adaptation.}
    \label{fig:environment_collision}
    \vspace{-4mm}
\end{figure}
\subsection{Environment-Collision Avoidance}
\label{ExperimentsEnvironmentCollision}
For all environment-collision avoidance tasks, we set ${\mu = 0.5}$ and ${\delta = 0.02}$. The map using \emph{FIESTA} is built with the resolution of $10$ cm using the LiDAR sensor on the robot. As shown in \figref{fig:robotCollisionModeling}, the robot is enclosed by spheres with $\delta_{max} = 40$~cm, $10$~cm, $5$~cm, $5$~cm, and $10$~cm for the base, shoulder, upper arm, elbow plus forearm, and the LiDAR cage respectively. 

\paragraph*{Dynamic Obstacles} To showcase the efficiency of the algorithm, we present whole-body planning for dynamic obstacle avoidance. An end-effector target position is imposed while a cart is pushed around the robot. Presented in \figref{fig:environment_collision}, the legged manipulator is capable of keeping track of the target while moving away from the approaching cart. Additionally, the base roll is exploited for better arm extension.

\paragraph*{Door Opening} Ultimately, we test our framework in an autonomous door-opening scenario. To avoid collision with the door frame during interaction in \cite{JeanPierre}, we imposed a high penalty on the lateral motion of the base while passing through the door. Enhanced with environment-collision avoidance through SDF, we can now automate the procedure without additional heuristics. We command the robot to push the door open to an angle of $100^{o}$. As we can see in \figref{fig:environment_collision}, the robot starts in front of the handle and is close to one side of the door frame. By adapting its torso yaw, the robot can navigate through the passage safely without any collisions. It can also avoid the approaching person during the execution.\footnote{https://www.youtube.com/watch?v=m3rJWJVzYuY}

\subsection{Computation Benchmark}
\label{sec:benchmark}
For both self-collision and obstacle avoidance, we collect five runs of reaching specific end-effector targets to benchmark the whole-body collision avoidance behaviors. Each run lasts around 1000 MPC iterations, and the average MPC computation per iteration over all five runs is analyzed. All the results are normalized w.r.t. the blind case, which means no self-collision and no environment-collision avoidance.
\newline
\begin{figure}[t]
    \begin{subfigure}[t]{0.48\textwidth}
        \includegraphics[width=\textwidth]{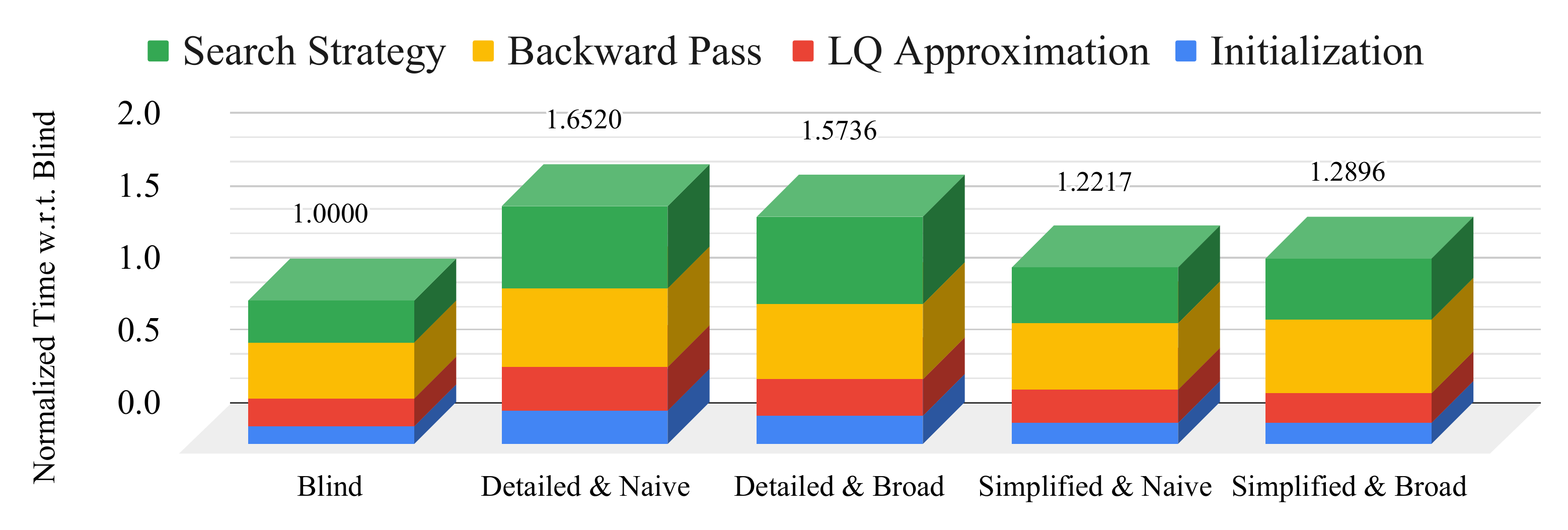}
    \end{subfigure}
    \label{fig:selfCollisionAvoidanceBenchmark}
    \begin{subfigure}[t]{0.24\textwidth}
        \includegraphics[width=\textwidth]{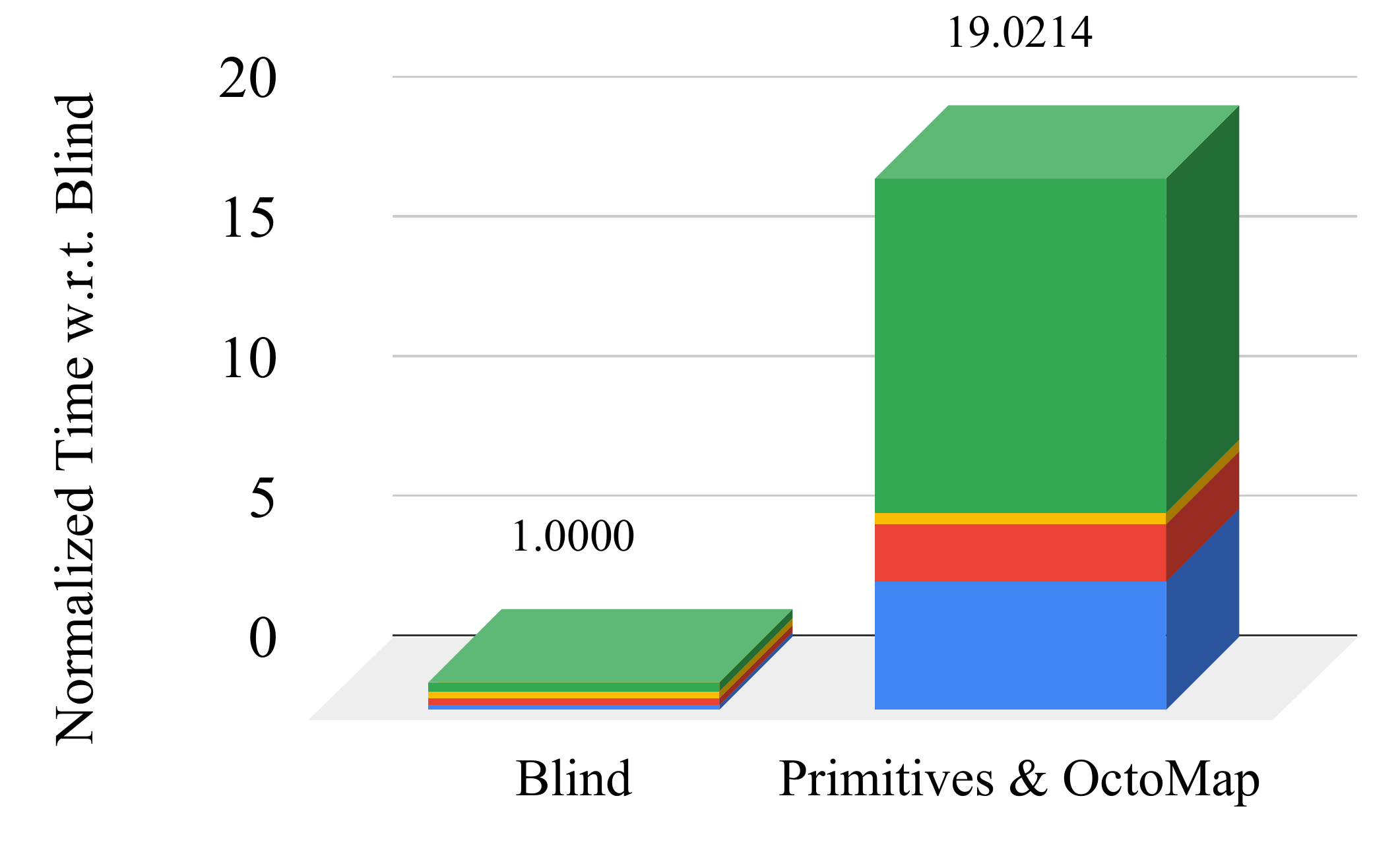}
    \end{subfigure}
    \begin{subfigure}[t]{0.24\textwidth}
        \includegraphics[width=\textwidth]{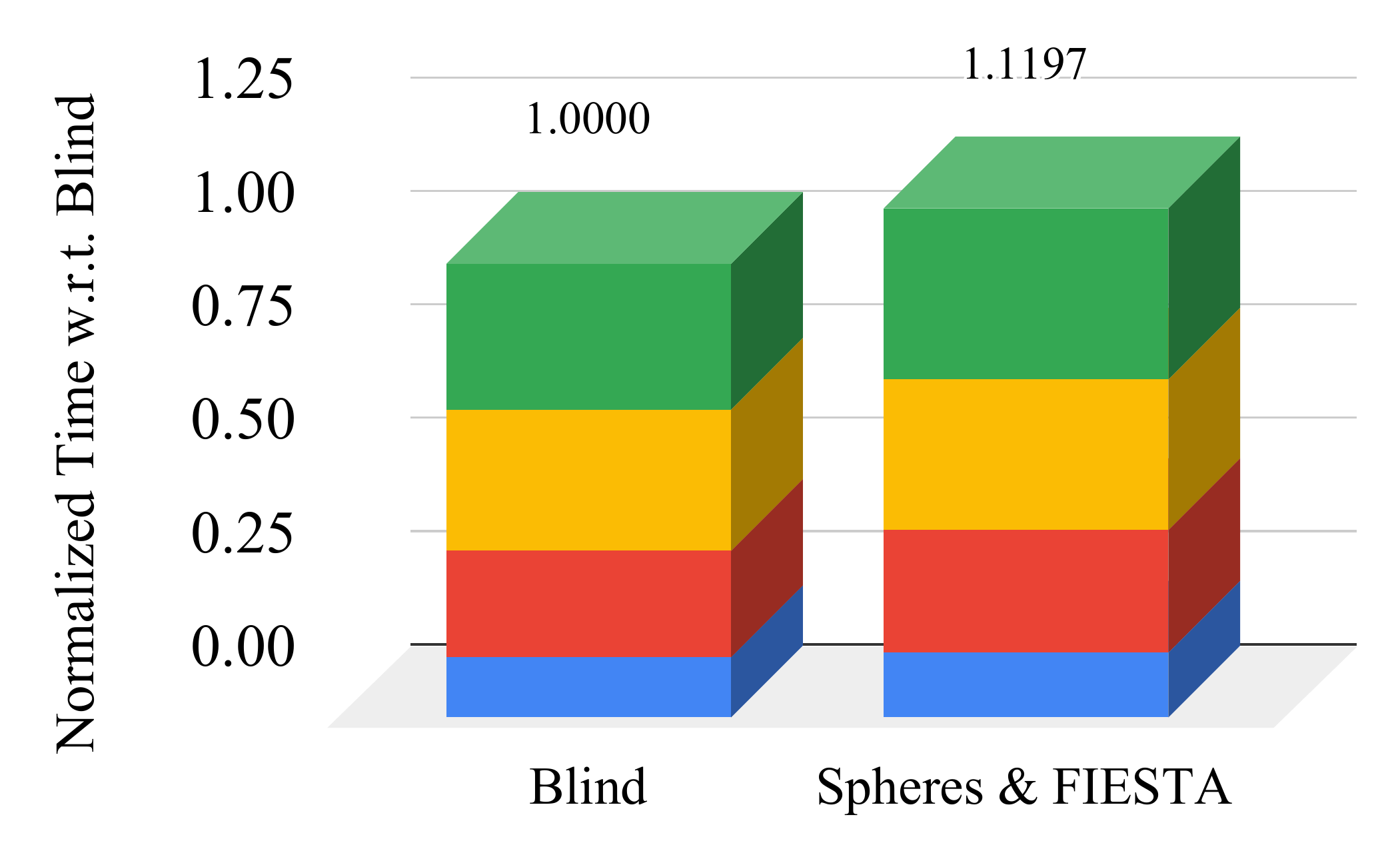}
    \end{subfigure}
    \caption{Average MPC computation per iteration for: (top) self-collision avoidance, (bottom left) environment-collision avoidance using collision primitives with \emph{OctoMap}, and (bottom right) using sphere approximations with \emph{FIESTA}.}
    \label{fig:benchmark}
    \vspace{-6mm}
\end{figure}

\subsubsection{Self-Collision Avoidance}
\label{BenchmarkSelfCollision}
\begin{figure}[t]
   \centering
   \includegraphics[width=0.25\textwidth, trim=0 0 0 6cm, clip]{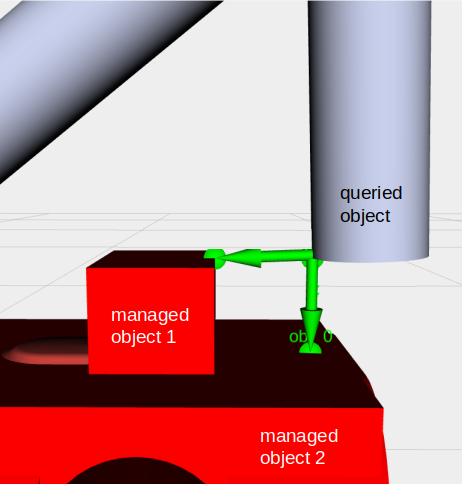}
   \caption{Direction discontinuity in gradients. The broad-phase technique may experience non-smooth jumps of distance gradient directions at certain configurations.}
   \label{fig:gradientIssue}
   \vspace{-4mm}
\end{figure}

We compare the four strategies against the blind case as shown in \figref{fig:benchmark}. As expected, using the detailed model with the na{\"i}ve approach is the most expensive. However, compared to the na{\"i}ve strategy, applying the broad-phase technique saves little time with the detailed model but is more expensive than the simplified model. For both models, broad-phase managing indeed speeds up the Linear Quadratic (LQ) Approximation step, which means less time for distance queries and gradient computation. Nevertheless, it takes longer to perform the line search step of the algorithm. The problem is that, at certain configurations, a manager may have multiple objects simultaneously closest to the queried object as illustrated in \figref{fig:gradientIssue}. This would result in a discontinuity of gradient directions, which is unfavorable for gradient-based optimization. Besides, as the broad-phase manager with the simplified model only saves about two narrow-phase queries, it reduces about $2\%$ in the LQ Approximation compared to the na{\"ive} approach. As a consequence, we opt for the na{\"i}ve strategy with the simplified robot collision model.
\subsubsection{Environment-Collision Avoidance}
\label{BenchmarkEnvironmentCollision}
To benchmark our method which relies on the primitive collision bodies, we also compare it to the proximity query introduced in \cite{PointCloudCollision} for distances between shape primitives and an \emph{OctoMap} \cite{OctoMap}. We build both the \emph{FIESTA} map and \emph{OctoMap} from the ``Cow and Lady dataset" in \cite{Voxblox} with 5 cm resolution. \figref{fig:benchmark} shows that using shape primitives with \emph{OctoMap} is 19 times as expensive as the blind case. This method essentially regards an \emph{OctoMap} as a broad-phase manager with each map cell as a collision box. Therefore, it would suffer from the gradient issue as well. The large amount of cells in the map is also the culprit of the extremely expensive computation. On the contrary, collision avoidance by reading the cached distances and gradients from the \emph{FIESTA} map given the sphere centers leads to only about $11.97\%$ increase in the computation time.

\section{DISCUSSION \& CONCLUSION}
\label{Conclusion}
In this paper, we endow the unified whole-body MPC framework \cite{JeanPierre} with self-collision and environment-collision avoidance capabilities, which enables dynamic maneuvers involving potential collisions to be safely executed on hardware. The robot collision model is represented with collision primitives. The na{\"i}ve self-collision distance query with the simplified model has proven its efficiency in real-time planning. By further enclosing the collision primitives with a set of collision spheres, fast updates of the \emph{FIESTA} map and computationally negligible queries of the pre-computed distances and gradients allow both static and dynamic obstacle avoidance. We conduct weight throwing and locomotion balancing with the swinging arm to demonstrate rapid whole-body self-collision avoidance during both free-motions and object manipulation scenarios. The autonomous door opening task in which the robot avoids the static door frame and the approaching human further validates the obstacle avoidance capabilities of the MPC framework. 

It is worth noting that although soft constraints cannot guarantee strict constraint satisfaction, allowing small violations facilitates the solver's convergence and prevents it from failing. Thus, the minimum allowed distance threshold for both collision constraints is used to provide room for constraint violations before the actual collision occurs. Moreover, environment-collision avoidance actually suffers from limitations of the LiDAR. For instance, the robot cannot see points in close proximity to the LiDAR. The arm in the LiDAR's field of view also creates a blind spot in the front. Furthermore, since the map update rate is dictated by the LiDAR update rate at $15$ Hz, the speed of the dynamic obstacles was limited during our experiments so that they don't suddenly appear within the distance threshold, thus blowing up the cost. Nonetheless, all these shortcomings can be alleviated with additional sensors.



\addtolength{\textheight}{0cm}   



\clearpage
\bibliographystyle{IEEEtran}
\bibliography{bibliography}

\end{document}